\documentclass[preprint,12pt]{elsarticle}

\usepackage{amssymb}

\journal{SIAM Journal on Imaging Sciences}
\usepackage{latexsym,bm,amsmath,amssymb}
\usepackage{enumerate}
\usepackage{graphicx}
\usepackage[ruled,vlined]{algorithm2e}
\usepackage{algorithmic}
\usepackage{amsfonts,amssymb,amsmath}
\usepackage{dsfont}
\usepackage{multirow}
\usepackage[table]{xcolor}
\usepackage{colortbl}
\usepackage[T1]{fontenc}
\usepackage{bm}
\usepackage{setspace}
\usepackage{caption}
\usepackage{color}

\usepackage{graphicx}
\usepackage{setspace}
\usepackage{float}
\usepackage{epsfig,epstopdf}
\usepackage{verbatim}
\usepackage{algorithmic}
\usepackage{array}
\usepackage{amsthm}
\usepackage{booktabs}
\usepackage{graphicx}
\usepackage{subfigure}
\usepackage{multirow}
\usepackage{threeparttable}
\usepackage{hyperref}
\usepackage{makecell}
\usepackage{fmtcount}
\usepackage{booktabs}

\usepackage{nicefrac}
\usepackage{xcolor}
\usepackage{xspace}
\usepackage{enumerate}
\usepackage{mathrsfs}



\newcolumntype{L}[1]{>{\raggedright\let\newline\\\arraybackslash\hspace{0pt}}m{#1}}
\newcolumntype{C}[1]{>{\centering\let\newline\\\arraybackslash\hspace{0pt}}m{#1}}
\newcolumntype{R}[1]{>{\raggedleft\let\newline\\\arraybackslash\hspace{0pt}}m{#1}}

\numberwithin{equation}{section}

\theoremstyle{remark}

\newcommand{\eg}{e.g.}
\newcommand{\ie}{i.e.}

\newcommand{\sanhao}{\fontsize{7pt}{\baselineskip}\selectfont}

\newcommand{\cI}{\mathcal{I}}

\newcommand{\vt}{\vartheta}

\newcommand{\R}{\mathbb{R}}

\newcommand{\norm}[2][]{\|{#2}\|_{#1}}

\newcommand{\diag}{\mathrm{diag}}

\newcommand{\scal}[2]{\left\langle #1,#2 \right\rangle}

\newcommand{\suml}[2]{\sum\nolimits_{#1}^{#2}}

\graphicspath{{figures/}}

\begin{document}
	
\begin{frontmatter}


\title{Poisson Noise Reduction with Higher-order\\ Natural Image Prior Model}


\author[labelFWS]{Wensen Feng}
\author[labelXY]{Hong Qiao}
\author[labelYJ]{Yunjin Chen\corref{cor3}}
\ead{chenyunjin$\_$nudt@hotmail.com}

\address[labelFWS]{School of Automation and Electrical Engineering, University of Science and Technology Beijing, Beijing 100083, China.}
\address[labelXY]{State Key Laboratory of Management and Control for Complex Systems, Institute of Automation, Chinese Academy of Sciences, Beijing 100190, China.}
\address[labelYJ]{Institute for Computer Graphics and Vision,
Graz University of Technology, 8010 Graz, Austria}
\cortext[cor3]{Corresponding author.}

\begin{abstract}
Poisson denoising is an essential issue for various imaging applications,
such as night vision, medical imaging and microscopy.
State-of-the-art approaches are clearly dominated by patch-based non-local
methods in recent years.
In this paper, we aim to propose a local Poisson denoising model with both
structure simplicity and good performance.
To this end, we consider a variational modeling to integrate the so-called
Fields of Experts (FoE) image prior, that has proven
an effective higher-order Markov Random Fields (MRF) model for many
classic image restoration problems. {\color{black}
We exploit several feasible variational
variants for this task. We start with a direct modeling in the original image domain
by taking into account the Poisson noise statistics, which performs generally well
for the cases of high SNR. However, this strategy encounters problem
in cases of low SNR. Then we turn to an alternative modeling strategy by using the
Anscombe transform and Gaussian statistics derived data term.
We retrain the FoE prior model directly in the transform domain.
With the newly trained FoE model, }we end up with a local variational
model providing strongly competitive results against state-of-the-art
non-local approaches, meanwhile bearing the property of simple structure.
Furthermore, our proposed model comes along with an additional advantage, that the inference is very efficient as
it is well-suited for parallel computation on GPUs.
For images of size $512 \times 512$, our GPU implementation takes less than 1 second to produce state-of-the-art Poisson
denoising performance.
\end{abstract}


\end{frontmatter}

\section{Introduction}\label{Section1}

The degradation of the acquired signal by Poisson noise is a common problem that arises in applications such as
biomedical imaging, fluorescence microscopy and astronomy imaging, among many others \cite{Bertero2010, Rodrigues08, Berry05}.
Thereby, the removal of Poisson noise is of fundamental importance especially for further processing, such as image segmentation and
recognition. However, the Poisson noise is not additive and its strength depends on the image intensity, alluding to the fact that
Poisson denoising is generally intractable relative to the usual case of additive white Gaussian noise.

Up to now, there exist plenty of Poisson denoising algorithms in the literatures, see \cite{INAnscombe7},
\cite{INAnscombe8}, \cite{Figueiredo}, \cite{INAnscombe10} and the references therein for a survey. Roughly speaking,
existing Poisson denoising approaches can be divided into two main classes: (1) with variance-stabilizing transformation (VST)
and (2) without VST.

Usually, the approaches in the first class take three steps for Poisson denoising. (Step 1:) A
nonlinear VST such as Anscombe \cite{Anscombe1,Anscombe2} or Fisz \cite{Fisz} is applied to the input data. This produces data with approximate
additive Gaussian noise, which is signal-independent. The rationale of applying a VST to the input data is
to remove the signal-dependency property of Poisson noise.
(Step 2:) As the transformed image can be approximately treated as a noisy image with Gaussian noise of unitary variance,
Gaussian denoising algorithms are applicable to estimate a clean image for the transformed image.
(Step 3:) The underlying unknown noise-free image is obtained by applying an
inverse VST \cite{INAnscombe1, Lefkimmiatis, INAnscombe3, INAnscombe5} to the denoised data.

Since the problem of Gaussian noise removal is well studied, there are many candidates applicable for the second step. A typical
choice is the well-known BM3D algorithm \cite{BM3D}, with which the resulting Poisson denoising algorithm can lead to
state-of-the-art performance. In general, the VST based approaches work very well for relative large-count Poisson denoising problems.
However, its performance will dramatically decrease for cases of low-counts \cite{Salmon2}, especially for extremely
low-count cases \eg, images with $\text{peak} = 0.1$. This motivates the other class of Poisson denoising algorithms, which directly
investigate the statistics properties of the Poisson noise, other than the transformed noise.

To avoid the VST operation, several authors have investigated reconstruction algorithms that rely directly on the Poisson
statistics. In \cite{Willettpoisson}, a novel denoising algorithm in consideration of the Poisson
statistics is developed, which is originated from the multiscale likelihood framework proposed by Kolaczyk and Nowak \cite{Kolaczyk}. In \cite{Salmon2}, J. Salmon $et$ $al.$ derived a Gaussian mixture model (GMM) \cite{Yu1} based
approach relying directly on the Poisson noise properties. This approach combines an adaptation
of Principal Component Analysis (PCA) for Poisson
noise \cite{PoissonPCA1} and sparse Poisson intensity estimation methods \cite{SPIRAL-TAP}, whereby it achieves state-of-the-art results for particularly high noise levels.
Two versions are involved in this method: the non-local PCA (NLPCA) and the non-local
sparse PCA (NL-SPCA). Among them, the NL-SPCA adds an $\ell_1$ regularization term to the minimized objective and achieves a better recovery performance. The similar data-fidelity term
derived from Poisson noise statistics is also adopted in \cite{Figueiredo}, \cite{Le1}, \cite{Giryes}. Especially, the work in \cite{Giryes}
proposes a new scheme for Poisson denoising and achieves state-of-the-art result. It relies on the Poisson statistical model and uses a
dictionary learning strategy with a sparse coding algorithm.

\subsection{Our contribution}
When we have a closer look at state-of-the-art Poisson denoising approaches,
we find that such approaches
are clearly dominated by patch-based non-local methods, such as PCA based model
\cite{Salmon2} and BM3D based algorithm \cite{INAnscombe5}.
Furthermore, some of them (e.g., \cite{Salmon2}) achieve
good denoising performance at the cost of computation efforts.
A notable exception is the BM3D based algorithm \cite{INAnscombe5},
which meanwhile offers high efficiency,
thanks to the highly engineered and well refined BM3D algorithm.

The goal of this paper/our contribution is to propose a local model with
simple structure,
while simultaneously with strongly competitive denoising performance to
those state-of-the-art non-local models.
To this end, we investigate variational models to incorporate a local
image prior model called Fields of Experts (FoE) \cite{RothFOE},
which has proven highly effective for
Gaussian denoising \cite{ChenGCPR, chenTIP} and multiplicative noise
reduction \cite{SpeckleFOE}.

{\color{black}
We consider several possible ways to construct the corresponding variational models.
We start from a straightforward variational model with a data term directly
derived from the Poisson noise statistics.
While this model generally performs well in cases of high SNR, it encounters
problems in cases of low SNR due to too many zero-valued pixels.
Motivated by the BM3D based algorithm \cite{INAnscombe5}, we then resort to
the Anscombe transform based framework, together with a quadratic data term, which is
derived from Gaussian statistics.

With the Anscombe transform based variational model, we
directly train the FoE image prior model in the Anscombe transform domain
by using a loss-specific training scheme. In experiments, we find that the quadratic
data term does not perform very well for the cases of low SNR, due to the
imprecise stabilization and markedly skewed distribution of the transformed data.
Therefore, we investigate an ad hoc data fidelity term for these circumstances
to improve the performance. }
Eventually, we reach a FoE prior based Poisson denoising model,
which can lead to competitive
performance against state-of-the-art algorithms,
meanwhile preserve the structure simplicity.

{\color{black}
In summary, our contribution comes from a successful adaptation of the FoE image
prior model for the problem of Poisson denoising via a general variational model,
which is capable of handling Poisson noise of any level.}
\subsection{Organization}
{\color{black}
The remainder of the paper is organized as follows. Section II presents brief reviews
for the building blocks exploited in our work.
}In the subsequent section III, we propose variant variational models to
incorporate FoE priors for the task of Poisson denoising.
Subsequently, Section IV describes comprehensive experiment results for the
proposed model.
The concluding remarks are drawn in the final Section V.

\section{Preliminaries}
\label{Preliminaries}
To make the paper self-contained, in this section we provide a brief review of
Poisson noise, the Anscombe Transform and FoE image prior models.
Due to the non-convexity of FoE regularization,
FoE related variational models impose non-convex optimization problems.
In order to efficiently solve these problem, we resort to a recently developed
non-convex optimization algorithm - iPiano \cite{iPiano}. For the sake of
readability, we also
present the basic update rule of the iPiano algorithm in this section.
Furthermore, as we repeatedly make use of the loss-specific learning scheme to train a
FoE regularizer for the Poison denoising task, in this section we also
present the general training process, which is treated as
a toolbox to train a FoE model.
\subsection{Poisson noise}
{\color{black}
Let $x\in \mathbb{R}^N$ be the original true image of interest and $y \in
{\mathbb{Z}_+^N}$ be the observed Poisson noisy image,
where the entries in $y$ (given $x$) are Poisson
distributed independent random variables with mean and variance $x_i$, i.e.,
\begin{equation}
P\left( y_i| x_i\right) = \left\{
\begin{array}{ll}
{\frac{{x_i}^{y_i}}{y_i !}\mathrm{exp}\left( - x_i\right)}, & \mbox{$x_i>0$}\\
\delta_0\left(y_i \right), & \mbox{$x_i = 0,$}
\end{array} \right.
\label{poissondistri}
\end{equation}
where $x_i$ and $y_i$ are the $i$-th component in $x$ and $y$ respectively, and $\delta_0$ is
the Kronecker delta function.
According to the property of the Poisson distribution, for the Poisson noise
$n_i = y_i - x_i$, we have
$\text{E}(n_i | x_i) = 0$ and $\text{var}(n_i | x_i) = x_i$. This alludes to
the fact that Poisson noise is signal dependent,
as the variance (strength) of the noise is proportional to the
signal intensity $x_i$. }Moreover, as the standard deviation of the noise $n_i$ equals $\sqrt{x_i}$, the signal-to-noise ratio (SNR)
is given as $\sqrt{x_i}$. Then the effect of Poisson noise
increases (\ie, the SNR decreases) as the intensity value
$x_i$ decreases.
Therefore, it is reasonable to define the noise level in the image $x$
by the maximal value (the peak value) in $x$, since lower intensity in the image yields a stronger noise.

According to the Gibbs function, the likelihood \eqref{poissondistri} leads to the following energy term
(also known as data term in the variational framework) via the equation $E = -\text{log}P(y|x)$
\begin{equation}
\langle x-y \mathrm{log}x,1 \rangle,
\label{poissonfidelity}
\end{equation}
where $\langle,\rangle$ denotes the standard inner product.
This data-fidelity term (\ref{poissonfidelity}) is the so-called Csisz{\'a}r I-divergence model \cite{csiszar1991least}, and
has been widely investigated in previous Poisson denoising algorithms, \eg, \cite{Salmon2,SPIRAL-TAP,Lingenfelter}.

\subsection{Variance stabilization with the Anscombe transform}
The Anscombe transform \cite{Anscombe1,Anscombe2} is defined as
\begin{equation}
v=f_{Anscombe}\left( y \right)=2\sqrt{y+\frac{3}{8}},
\label{Anscombetransform}
\end{equation}
{\color{black}Applying this nonlinear
transform to Poisson distributed data gives a signal with approximate additive
Gaussian noise of unitary variance.
Concerning the inverse transform, \cite{INAnscombe5}
exploited three types of optimal inverse Anscombe transforms, which
can significantly improve the recovery performance obtained by the direct
algebraic inverse, particularly in the circumstances of the low-count. }
In this paper we consider the following closed-form approximation of
the exact unbiased inverse of the Anscombe transform
\cite{makitalo2011closed}
\begin{equation}
\label{invAns}
\cI_C(z) = \frac 1 4 z^2 + \frac 1 4 \sqrt{\frac 3 2 } z^{-1}- \frac {11}{8} z^{-2} + \frac 5 8 \sqrt{\frac 3 2} z^{-3} - \frac 1 8 \,.
\end{equation}

\subsection{The FoE image prior model}
{\color{black}
The FoE image prior model \cite{RothFOE} is defined by a set of linear filters
and a potential function.
This image prior model has been widely investigated for many classic image restoration problems due to its effectiveness
\cite{chenTIP, SamuelMRF, GaoCVPR2010}.
Given an image $w$, the FoE prior model is formulated as the following
energy functional
\begin{equation}\label{FOEmodel}
E_{FoE}(w) =
\suml{i=1}{N_f}{\alpha_i} \suml{p=1}{N} \rho((k_i * w)_p),
\end{equation}
}where $N_f$ is the number of filters, $k_i$ is a set of learned linear filters with the corresponding weights $\alpha_i > 0$, $N$ is the number of pixels in image $w$,
$k_i * w$ denotes the convolution of the filter $k_i$ with image $w$,
and $\rho(\cdot)$ denotes the potential function. It is clear that
the FoE prior model is defined upon locally supported filter responses,
oriented towards local directions and textural features.

In previous works, the potential function is usually chosen as a non-convex function.
In this paper, we follow the common choice of the potential function $\rho$, defined as
\begin{equation}
\rho(z) = \text{log}(1 + z^2)\,.
\label{STfunc}
\end{equation}

\subsection{The iPiano algorithm}
The iPiano algorithm is designed for solving an optimization problem which is composed of
a smooth (possibly non-convex) function $F$ and a convex (possibly non-smooth) function $G$:
\begin{equation}\label{fplusg}
\arg\min\limits_{u}F(u) + G(u) \,.
\end{equation}
It is based on a forward-backward splitting scheme with an inertial force term. Its basic update
rule is given as
\begin{equation}\label{iPianoupdate}
u^{n+1}=\left( I+\tau \partial G \right)^{-1}\left( u^n-\tau \nabla F\left( u^n \right) +
\gamma \left( u^n - u^{n-1} \right) \right),
\end{equation}
where $\tau$ and $\gamma$ are the step size parameters.
$u^n-\tau \nabla F\left( u^n \right)$ is the forward gradient descent step,
and $\gamma \left( u^n - u^{n-1} \right)$ is the inertial term.
The term $\left( I+\tau \partial G \right)^{-1}$ denotes the
standard proximal mapping \cite{nesterov2004introductory}, which is the backward step.
The proximal mapping $\left( I+\tau \partial G \right)^{-1}(\tilde u)$ is given as the following
minimization problem
\begin{equation}\label{subproblemG}
\left( I + \tau \partial G \right)^{-1}(\tilde{u}) = \arg\min\limits_{u} \frac{\|u - \tilde
{u}\|^2_2}{2} + \tau G(u)\,.
\end{equation}

\subsection{Loss-specific learning scheme to train the FoE prior model}
\label{LossTraining}
In our work we exploit a loss-specific training scheme to learn FoE prior models
for the Poisson denoising problem. Here we first
summarize the general training framework, and in latter sections
we repeatedly use it as a toolbox to train FoE prior models.

\subsubsection{Loss-specific training model for the FoE prior}
{\color{black}Given a set of clean/noisy image pairs $\{g_s, f_s\}_{s=1}^{N_s}$ where $g_s$ and $f_s$ are the $s^{th}$ clean image and the associated noisy
image corrupted by Poisson noise, respectively, the loss-specific training scheme
is to learn optimized parameters $\vt = (\alpha, \beta)$ to minimize certain
predefined loss function. In summary, the learning model is formulated as
the following bilevel optimization problem
\begin{equation}\label{learningmodel}
\begin{cases}
\min\limits_{\vt}L(w^*(\vt)) =
\suml{s=1}{N_s} \ell(w_s^*(\vt), g_s)\\
\text{s.t.}~
w_s^*(\vt) = \arg\min\limits_{w} E(w) =
\suml{i=1}{N_f}e^{\alpha_i} \suml{p=1}{N} \rho((k_i * w)_p) + D(w, f_s)\,,
\end{cases}
\end{equation}
where the lower-level problem is defined as the FoE prior regularized
variational model, and the upper level problem is defined by a loss
function $\ell(w^*(\vt), g)$ to measure the difference between the optimal
solution of the lower-level problem $w^*$ and the ground-truth $g$.
At present, we do not specify the specific form of the data term $D(w, f)$ in the
lower-level problem, as it can have different forms in our work.

In the training phase, the FoE model is parameterized by $\alpha$, which is
related to the weights of the filters, and by $\beta$, which is related to
the filters coefficients. we use $e^{\alpha_i}$ instead of $\alpha_i$ to ensure
the learned weights are positive. The dependency of $k_i$ on $\beta_i$ is defined as
\begin{equation}
\label{filtercoe}
k_i = \suml{j=1}{N_b}\beta_{ij}b_j = B \beta_i\,,
\end{equation}
where $\beta_i = (\beta_{i1}, \beta_{i2},\cdots,\beta_{iN_b})^\top$,
$B=\{b_1,\cdots, b_{N_b}\}$ are usually chosen as a set of zero-mean
basis filters to keep consistent with the findings in \cite{Huang1999_Statistics}
that meaningful filters should be zero-mean. }

\subsubsection{Optimization for the training phase}
Previous works \cite{chenTIP,SamuelMRF} have demonstrated that it is possible to
find a meaningful stationary point via gradient-based optimization algorithms for
the learning problem \eqref{learningmodel}.
The gradient of the loss function with respect to the parameters $\vt$ can be
computed via implicit differentiation
\cite{SamuelMRF}
\begin{equation}\label{implicitdifferentiation}
\frac {\partial L}{\partial \vt} = - \frac {\partial^2 E}{\partial w \partial \vt}
\left( \frac {\partial^2 E}{\partial {w}^2}\right)^{-1}
\frac {\partial L}{\partial w} \bigg|_{w = w^*}
\end{equation}

In \eqref{implicitdifferentiation}, $\frac {\partial^2 E}{\partial {w}^2}$ denotes the Hessian matrix of $E(w)$,
\[
\frac {\partial^2 E}{\partial {w}^2} = H_E(w) = \suml{i=1}{N_f}e^{\alpha_i} K_i^\top \Gamma_i K_i +
\frac {\partial^2 D}{\partial {w}^2} \,,
\]
where $K_i$ is an $N \times N$ highly sparse matrix, implemented as 2D convolution of the image $w$ with filter kernel $k_i$,
\ie, $K_i w\Leftrightarrow k_i * w$, and matrix $\Gamma_i \in \R^{N \times N}$ is given as
\[
\Gamma_i = \diag (\rho''((K_i w)_1),\cdots,\rho''((K_i w)_N))
\]
with
\[\rho''(z) = \frac{2(1-z^2)}{(1+z^2)^2} \,.
\]
\begin{algorithm*}\caption{Generalized loss-specific training scheme for the FoE prior}\label{algo1}
\textbf{Input}: Training samples $\{g_s,f_s\}_{s=1}^{N_s}$ \\
\textbf{Output}: Learned filters $k_i$ with associated weights $e^{\alpha_i}$
\begin{itemize}
\item[1.]
Initialize parameters with $\vt^0 = \{\alpha^0, \beta^0\}$, let $n = 0$
\item[2.] For each training sample, minimize the energy functional
\begin{equation*}
\suml{i=1}{N_f}e^{\alpha^n_i} \suml{p=1}{N} \rho((K_i w)_p) + D(w, f_s)\,, ~(k_i = \suml{j=1}
{N_b}\beta^n_{ij}b_j),
\end{equation*}
to obtain ${w}_s^*(\vt^n)$
\item[3.] Compute $\frac {\partial L}{\partial \vt}$ at $\vt^n$ using \eqref{overallderivative} and the corresponding
loss function value $L$
\item[4.] Update parameters $\vt = \{\alpha, \beta\}$ using L-BFGS, let $n = n + 1$, and goto (2)
\end{itemize}
\end{algorithm*}
Then we arrive at the following formulations to compute the gradients
\begin{equation}\label{overallderivative}
\begin{cases}
\frac {\partial L}{\partial \beta_{ij}} =
- e^{\alpha_i}(B_j^T\rho'(K_i w) + K_i^\top \Gamma_i B_j w)^\top \cdot \left(H_E(w)\right)^{-1} \cdot \frac
{\partial L}{\partial {w}}
\\
\frac {\partial L}{\partial \alpha_{i}} =  - e^{\alpha_i}(K_i^\top \rho'(K_i w))^\top \cdot \left(H_E(w)\right)^{-1} \cdot \frac {\partial L}
{\partial {w}} \,,
\end{cases}
\end{equation}
where $\rho'(K_i w) = (\rho'((K_i w)_1),\cdots,\rho'((K_i w)_N))^\top \in \R^{N}$,
with
\[
\rho'(z) = \frac {2z}{1+z^2}\,.
\]

Note that all the expressions in \eqref{implicitdifferentiation} and \eqref{overallderivative}
are evaluated at ${w}^*(\vt)$, and therefore we first have to
solve the lower-level problem for a given parameter vector $\vt$ before computing the gradient of $L(\vt)$.
Equation \eqref{overallderivative} is the derivatives for the case of
one training sample. Considering the case of $N_s$ training samples,
the derivatives of the overall loss function with respect to the parameters $\vt$ are just the sum of
\eqref{overallderivative} over the training dataset.

Now we can exploit gradient-based algorithm for optimization. In this work, we employed
a quasi-Newton's method - L-BFGS \cite{BFGS} for optimization.
Let us summarize our training algorithm in Algorithm~\ref{algo1}. Note that the step (2) in Algorithm~\ref{algo1}
is completed using the iPiano algorithm. More details will be described in Section \ref{section3}.
The training algorithm L-BFGS is terminated when the relative change of the loss is less than a tolerance, e.g., $tol = 10^{-5}$, a maximum number of iterations e.g.,
$maxiter = 500$ is reached or L-BFGS can not find a feasible step to decrease the loss.

\subsection{Details of the training setup for Poisson denoising}
It is shown in \cite{chenTIP} that, although the loss specific training scheme is a
discriminative training approach, the trained FoE image prior model is not
highly-tailored to the noise level used for training, but in contrast,
it generalizes well for different noise levels. \textit{
Therefore, in the circumstance of Poisson denoising, we also only train the FoE image prior model once at a certain noise level,
and then apply it for different noise levels.}

Taking into account the fact that generic natural image priors have an interesting effect only at medium noise levels
\cite{levin2011natural}, we conduct our
training experiments based on the images with relatively high peak value. In this study, we choose $\text{peak} = 40$.

As we focus on the problem of Poisson denoising for natural images in a general sense,
FoE prior models should be trained based on natural image datasets.
We conducted our training experiments using the training images from the BSDS300 image
segmentation database~\cite{amfm2011}.
We used the whole 200 training images, and randomly sampled one $128 \times 128$
patch from each training image,
resulting in a total of 200 training samples. The gray value range of all the
images was first
scaled with peak value $\text{peak} = 40$, and then corrupted with
Poisson noise using the Matlab function
\textit{poissrnd}.

Concerning the model capacity of the trained FoE models,
following a common setup of filters for the FoE model in \cite{ChenGCPR},
we learned 48 filters with dimension $7 \times 7$. As we focused on mean-zero filters,
we made use of a modified DCT-7 basis (the atom with constant entries is excluded) to construct
our learned filters. We initialized the filters using the modified DCT-7 basis.
All filters have the same norm and weight, which are 0.1 and 1, respectively\footnote{
As shown in previous work \cite{ChenGCPR}, the training process is not sensitive to the initialization.}.

{\color{black}
As shown in the Algorithm~\ref{algo1}, the main computation burden
in the training phase is that: for each training sample, we need to solve the
lower-level with iPiano, and then calculate the gradients with respect to
parameters $\vt$
according to \eqref{overallderivative}, where a matrix inverse operation is involved.
With our unoptimized Matlab implementation, the training time for the setup considered
in this work (48 filters of size $7\times 7$) was approximately 24 hours on CPUs
Intel X5675, 3.07GHz.}

\section{Proposed Variational Models for Poisson Noise Reduction}
\label{section3}
In this section, we propose several variational models to incorporate FoE priors
for the task of Poisson denoising.
As we will investigate two different FoE prior models, we make use of
the following notations for clarity
\begin{itemize}
\item[$\textbf{FoE}_{\textbf{s}}$] trained in the original image domain with the
Poisson noise statistics derived data term;
\item[$\textbf{FoE}_{\textbf{A}}$] trained in the Anscombe transform domain with a quadratic data term.
\end{itemize}

\subsection{A straightforward formulation}
We started with a straightforward formulation to integrate a FoE prior for
Poisson denoising. It is given as the following
energy minimization problem
\begin{equation}\label{straightforward}
\color{black}
\arg\min\limits_{x > 0}E(x) =
\suml{i=1}{N_f}{\alpha_i}\suml{p=1}{N} \rho((k_i * x)_p) + \lambda \langle x-y \mathrm{log}x,1 \rangle \,,
\end{equation}
where $\lambda$ is a trade-off parameter.
This model is a direct combination of the FoE prior model \eqref{FOEmodel} and
the Poisson noise statistics derived data term \eqref{poissonfidelity}.
Regarding the FoE prior model, we exploited the loss-specific training scheme
described in Section \ref{LossTraining}, and
the lower-level problem is defined as the above Poisson denoising model,
i.e., we have the data term
\[
D(x, f) = \lambda \langle x - f \mathrm{log}x,1 \rangle \,,
\]
with $f = y$ where $y$ is the noisy input.
With respect to the upper-level problem, we considered the quadratic loss function
\[
\ell(x, g) = \frac 1 2\|x - g\|^2_2 \,,
\]
with $g$ denoting the ground-truth (clean) image. Then it is easy to check that
\[
\frac{\partial^2 D}{\partial x^2} = \text{diag}\left(\frac{f_1}{x_1^2}, \cdots, \frac{f_N}{x_N^2}
\right) \,,
\mathrm{and} \,\,
\frac{\partial \ell}{\partial x} = x - g\,,
\]
which are required to compute the gradients in \eqref{overallderivative}. Now we
can run training experiments based on the Poisson denoising model using the
Algorithm \ref{algo1}. Then we obtain a specialized FoE model
prior $\textbf{FoE}_{\textbf{s}}$, which is
directly trained based on the Poisson denoising model \eqref{straightforward}.
The learned filters are shown in Figure \ref{filters}(a).

The corresponding minimization problem \eqref{straightforward} with $\textbf{FoE}_{\textbf{s}}$ can be solved by the iPiano algorithm. Casting \eqref{straightforward}
in the form of \eqref{fplusg}, we have $F(x) = \suml{i=1}{N_f}{\alpha_i} \rho(k_i * x)$ and
$G(x) = \lambda \langle x-y \mathrm{log}x,1 \rangle$. It is easy to check that
\begin{equation}\label{FoEF}
\nabla_x F(x)= \suml{i=1}{N_f}{\alpha_i} K_i^\top \rho'({K_i} x) \,,
\end{equation}
and the proximal mapping with respect to $G$ is given by the following point-wise operation
\begin{equation}\label{subproblemGIdiv}
\left( I + \tau \partial G \right)^{-1}(\tilde{x})  = \frac{ \tilde{x} - \tau \lambda
+\sqrt{\left( \tilde{x} -\tau \lambda \right)^2+4\tau \lambda y}}{2} \,.
\end{equation}
The calculation rule \eqref{subproblemGIdiv}
for solving $x$ can guarantee that the constraint $x > 0$ is fulfilled when the input image $y$ is strictly positive.

\begin{figure*}[t!]
\centering
    \subfigure[{\tiny Noisy (19.50/0.3047)}]{\includegraphics[width=0.23\textwidth]{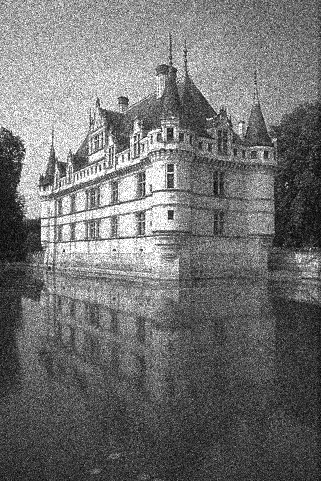}}\hfill
    \subfigure[{\scriptsize BM3D \cite{INAnscombe5}(28.89/0.8473)}]{\includegraphics[width=0.23\textwidth]{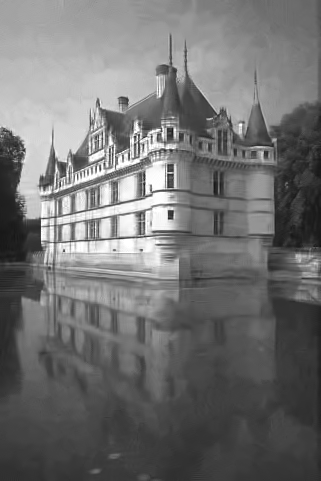}}\hfill
    \subfigure[{\scriptsize Model \eqref{straightforward} with $\textbf{FoE}_{\textbf{s}}$ (28.63/0.8405)}]{\includegraphics[width=0.23\textwidth]
{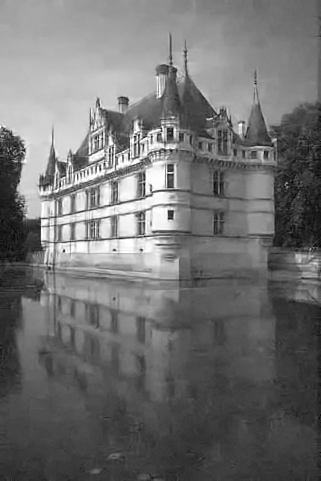}}\hfill
    \subfigure[{\scriptsize Model \eqref{overallModel} with $\textbf{FoE}_{\textbf{A}}$ (28.88/0.8481)}]{\includegraphics[width=0.23\textwidth]
{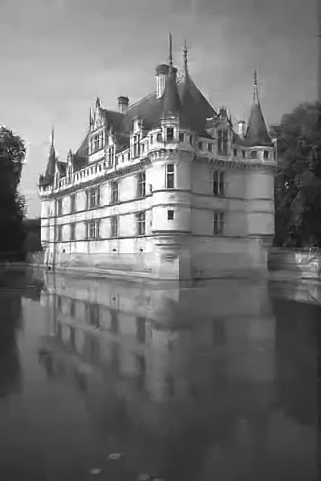}}\hfill
    \caption{\color{black}Poisson denoising example with $\text{peak} = 40$ on the image
``water-castle''. In this example, the peak value is relatively high. Both the
direct modeling \eqref{straightforward} and the modeling method in the Anscombe
transform \eqref{overallModel} can produce general good performance.
The results are reported by PSNR/MSSIM index.}
\label{poisson_preliminary}
\end{figure*}

We show a Poisson denoising example using model \eqref{straightforward}
with $\textbf{FoE}_{\textbf{s}}$ in Figure \ref{poisson_preliminary}(c).
We also present the result of the BM3D-based algorithm \cite{INAnscombe5} for
a comparison. A quick glance at the our result shows that the investigated model
\eqref{straightforward} achieves overall quite good result.

\begin{figure}[t!]
\centering
    \subfigure[{\tiny Noisy (9.32/0.1288)}]{\includegraphics[width=0.3\textwidth]{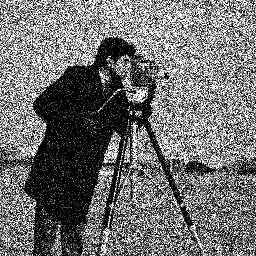}}
    \subfigure[{\scriptsize BM3D \cite{INAnscombe5} (23.96/0.7035)}]{\includegraphics[width=0.3\textwidth]
{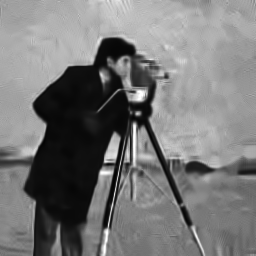}}
    \subfigure[{\scriptsize Model \eqref{lowcase} with $\textbf{FoE}_{\textbf{A}}$ (24.40/0.7519)}]{\includegraphics[width=0.3\textwidth]
{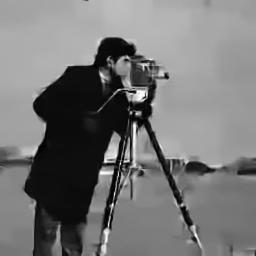}}\\
    \subfigure[{\scriptsize Model \eqref{straightforward} with $\textbf{FoE}_{\textbf{s}}$ and $c = 0$ (22.98/0.6418)}]{\includegraphics[width=0.3\textwidth]
{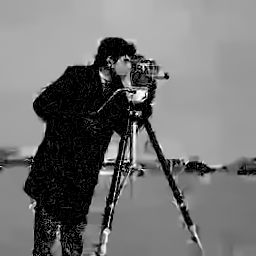}}
   \subfigure[{\scriptsize Model \eqref{straightforward} with $\textbf{FoE}_{\textbf{s}}$ and $c = 0.1$ (24.17/0.7331)}]{\includegraphics[width=0.3\textwidth]
{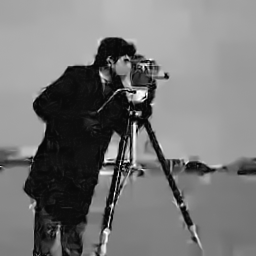}}
   \subfigure[{\scriptsize Model \eqref{straightforward} with $\textbf{FoE}_{\textbf{s}}$ and $c = 0.5$ (22.33/0.6751)}]{\includegraphics[width=0.3\textwidth]
{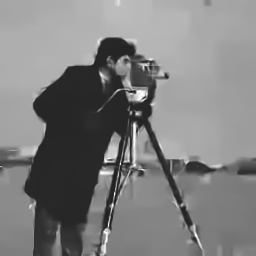}}
    \caption{\color{black}
Poisson denoising example with $\text{peak} = 4$ on the image ``cameraman''. The direct
model \eqref{straightforward} encounters problem
for this case of low peak in the dark regions. However, denoising in the
Anscombe transform can solve this problem.
The results are reported by PSNR/MSSIM index.}\label{lowpeak}
\end{figure}

\subsubsection{Problem of model \eqref{straightforward} for cases of low peak value}
In \eqref{straightforward}, it is clear that for the points of $y_p = 0$, the data term has the minimizer $x_p = 0$. In other words, if the points of $y_p = 0$ can not be filled by the prior term, they will not be updated and fixed at 0 in
the iterative process.
For the above exploited example with $\text{peak} = 40$, as only few pixels
have the gray value 0,
these points can be filled by the prior term.

{\color{black}
However, in cases of low peak value, large regions with many zeros frequently
appear in the image. Meanwhile, noisy pixels with very high intensity also appear
in these regions, and thus strong edges forms. As these regions are
significantly larger than the filter size of the FoE prior, the prior term
is not able to fill those zeros. As a consequence, those pixels with zero intensity
can not be updated and fixed at 0 in the iterations. Finally, these noisy
strong edges are retained, instead of smoothed out.
An example of $\text{peak} = 4$ is shown in Figure \ref{lowpeak}, where large
patches with many zeros appear
in the dark parts of the image, e.g., the coat of the cameraman. }

If we directly apply the model \eqref{straightforward} with the prior term $\textbf{FoE}_{\textbf{s}}$ to this problem,
the obtained result is shown in Figure \ref{lowpeak}(d).
As expected, there is a problem in those regions with many zeros.
We can remedy this problem by setting those zero-valued pixels in the noisy
images to certain constant $c$. The result with $c = 0.1$ is given
in Figure \ref{lowpeak}(e). It indeed improves the performance in the dark
regions, but it is still unsatisfactory. If we consider a
larger $c = 0.5$, it can indeed remove the artifacts in these dark regions,
but it brings a overall luminance bias, as such a
large $c$ changes the original input image too much.

\textit{
It turns out that the straightforward modeling \eqref{straightforward} is
not suitable for cases of low peak value, and this
motivates us to consider other alternatives for Poisson denoising. Inspired by the BM3D-based algorithm \cite{INAnscombe5}, which
perform Poisson denoising based on the Anscombe variance-stabilizing transformation, we also attempt to exploit this strategy, i.e., performing noise
reduction in the transform domain. Note that the points of $y_p = 0$ become
those of $v_p = \sqrt{\frac{3}{2}}$ after Anscombe transformation, thereby there is no the aforementioned problem in model \eqref{straightforward} for low peak cases.}

{
\begin{figure}[t!]
\centering
    \subfigure[The learned $\textbf{FoE}_{\textbf{s}}$ model]
{\includegraphics[width=0.48\textwidth]{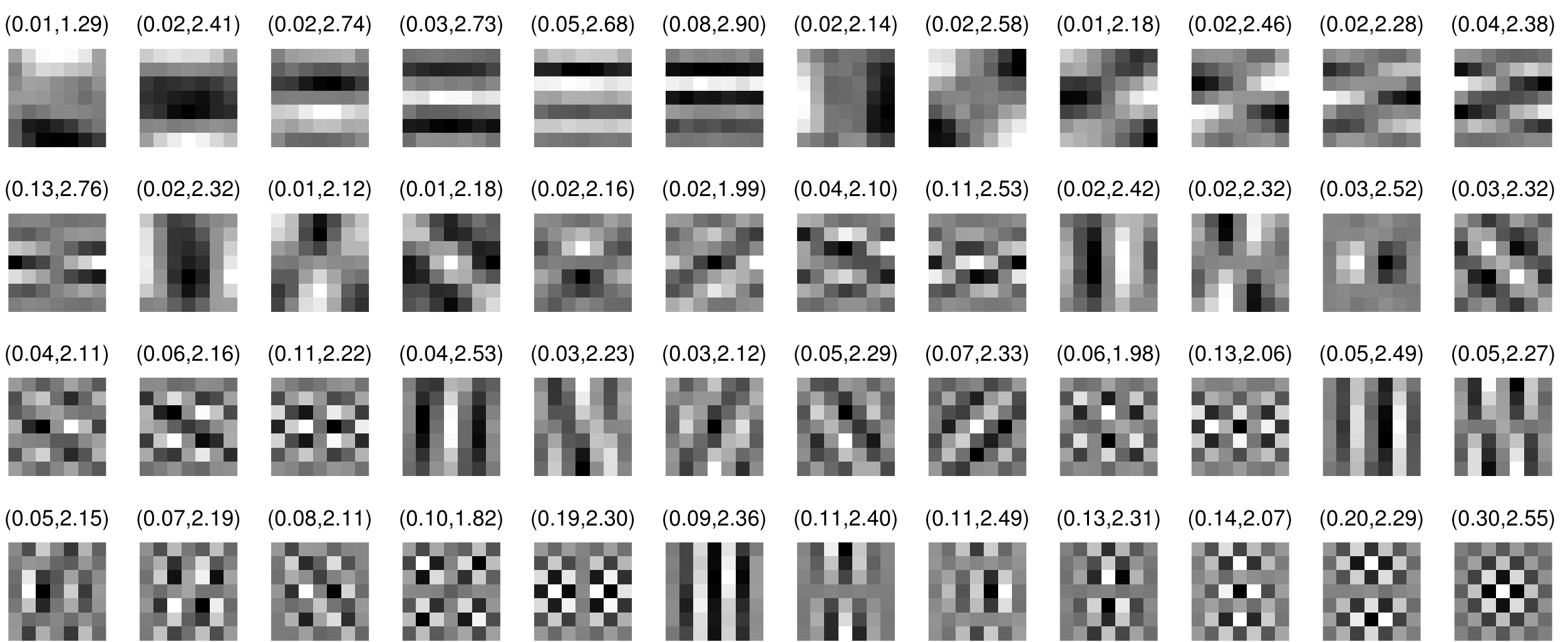}}
{\color{blue}\vrule height 15ex width 1.25pt }
	 \subfigure[The learned $\textbf{FoE}_{\textbf{A}}$ model]
{\includegraphics[width=0.48\textwidth]{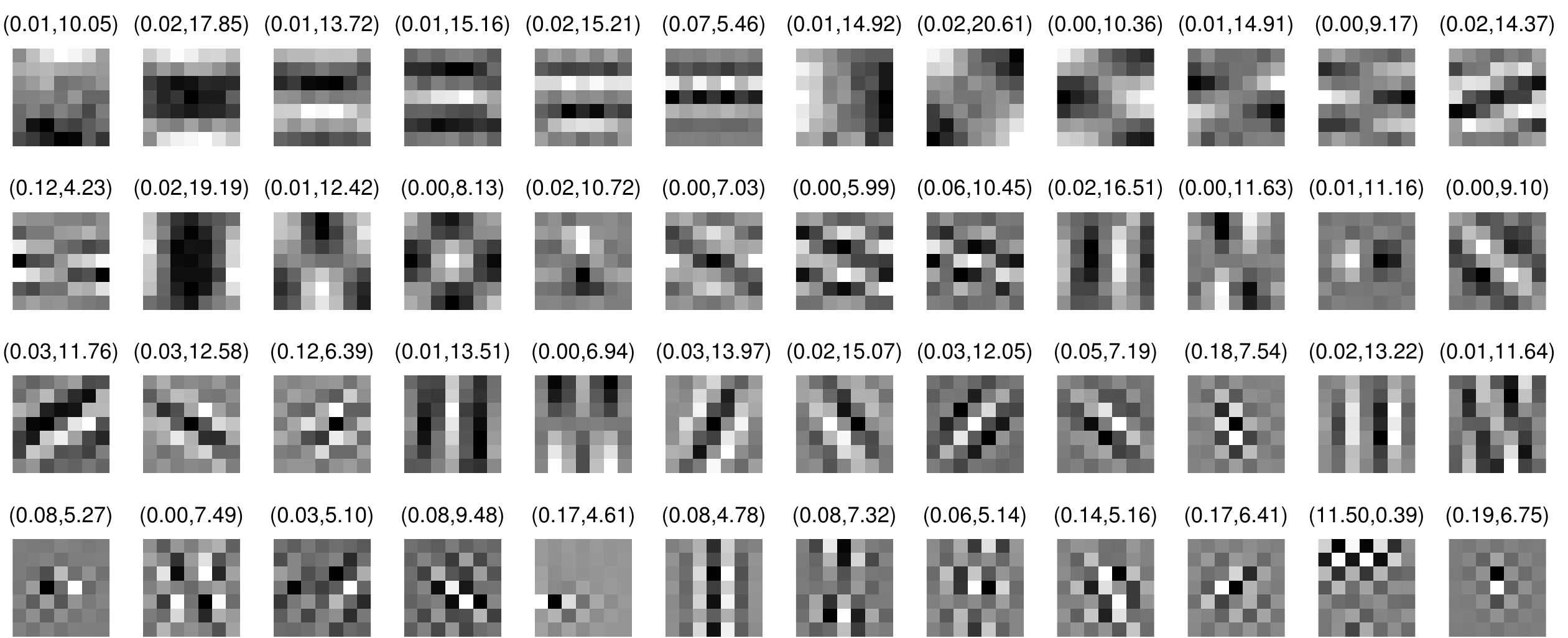}}
    \caption{\color{black} Two FoE prior models learned in this paper (
48 learned filters of size $7 \times 7$). The first number in the bracket is
the weight $\alpha_i$ and the second one is the norm of the filter. The learned FoE
prior models in the original image domain and the Anscombe transform domain
exhibit different appearances.}
\label{filters}
\end{figure}}

\subsection{Poisson denoising in the Anscombe transform domain}\label{AnscombePoisson}
Following the strategy of \cite{INAnscombe5}, we consider the following three-step procedure
to integrate the FoE prior model for Poisson denoising.
First, the noisy data $y$ is modified by applying the Anscombe transform in \eqref{Anscombetransform} to generate the data $v$.
The second step is to denoise the transformed data $v$ via a variational model
based on the trained FoE image
prior model. Finally, an estimate of the underlying image $x$ is obtained by applying the
inverse transform \eqref{invAns} to the denoised result $\hat{u}$. Therefore, the overall denoising procedure is given by
\begin{equation}\label{overallModel}
\begin{cases}
v = f_{Anscombe}\left( y \right) \,,\\
\hat{u} =\arg\min\limits_{u} \suml{i=1}{N_f}{\alpha_i} \suml{p=1}{N} \rho((k_i * u)_p) + \frac \lambda 2 \|u - v\|^2_2 \,,\\
\hat x = \cI_C(\hat{u}) \,.
\end{cases}
\end{equation}
The second step is the key point of the proposed Poisson denoising algorithm, where we make use of a quadratic data term as
the noise statistics of the transformed data can be approximately treated as additive Gaussian noise. The proximal mapping with
respect to this new data term is given by the following point-wise operation
\begin{equation}\label{subproblemGL2}
\left( I + \tau \partial G \right)^{-1}(\tilde u) = \frac{\tilde{u}+\tau \lambda  v
}{1+ \tau \lambda}\,,
\end{equation}
which is required by the iPiano algorithm.

{\color{black}
Note that the FoE prior model in \eqref{overallModel} performs in the Anscombe
root transform domain which will make a nonlinear change to the image dynamical
range. Therefore, we need to retrain the FoE model directly based on the
variational model in \eqref{overallModel}. Comparing two set of filters
learned in the original image domain and the Anscombe transform domain, respectively,
as shown in Figure \ref{filters}, one can see that have different appearances,
especially those filters with relatively high weights. }

We still make use the loss-specific training scheme described in Section \ref{Preliminaries}. Now, the lower-level problem is given as
\[
\color{black}
u^* =\arg\min\limits_{u} \suml{i=1}{N_f}e^{\alpha_i} \suml{p=1}{N} \rho((k_i * u)_p) + \frac 1 2 \|u - v\|^2_2 \,,
\]
where we omit the trade-off parameter $\lambda$ as it can incorporated into the weights $\alpha$. The corresponding upper-level
problem is defined as the quadratic loss function
\[
\ell (u^*, g) = \frac 1 2 \|\cI_C(u^*) - g\|_2^2 \,.
\]
It is easy to check that
\[
\frac{\partial^2 D}{\partial u^2} = I \,,
\]
and the gradient $\frac {\partial \ell}{\partial {u}}$ is given as
\[
\frac {\partial \ell}{\partial {u}} = \frac {\partial \cI_C(u)}{\partial {u}} \frac {\partial \ell}{\partial \cI_C(u)} =
\Lambda (\cI_C({u}) - g) \,,
\]
matrix $\Lambda \in \R^{N \times N}$ is given as
\[
\Lambda = \diag (\cI_C'(u_1),\cdots,\cI_C'(u_N))
\]
with $\cI_C'(z) = \frac 1 2 z - \frac 1 4 \sqrt{\frac 3 2 } z^{-2}+\frac {11}{4} z^{-3} - \frac {15} {8} \sqrt{\frac 3 2} z^{-4}$.
Then we can exploit the Algorithm \ref{algo1} to train FoE prior models in the
Anscombe transform domain. The obtained FoE prior model is
named as $\textbf{FoE}_{\textbf{A}}$.

{\color{black}
\subsection{An ad hoc data term for the cases of low peak in the Anscombe transform}
\label{Modeling}
For the denoising method in the  Anscombe transform domain, the corresponding
data term is usually chosen as a quadratic function, which is derived from
the Gaussian noise statistics. }Once we have obtained a new FoE prior
model $\textbf{FoE}_{\textbf{A}}$ for the Anscombe transform based data,
we can apply it for Poisson denoising with various noise levels\footnote{
Note that as the exploited FoE prior model is trained based on the case of $\text{peak} = 40$, we need to tune the trade-off
parameter $\lambda$ for other specific noise levels such that the result is optimized. It is effortless to tune the parameter $\lambda$,
and a detailed guideline will be presented in the implementation after acceptance. }
by using the framework \eqref{overallModel}.
An example for the case of $\text{peak} = 40$ is shown in Figure \ref{poisson_preliminary}(d). One can see that
in terms of both quantitative results and visual inspection, it achieves better results than those straightforward models, and moreover,
it leads to strongly competitive result to the BM3D-based
algorithm \cite{INAnscombe5}.
\begin{figure*}[t]
\centering
\subfigure[{\scriptsize Noisy image (peak = 2)}]{
\centering
\includegraphics[width=0.27\textwidth]{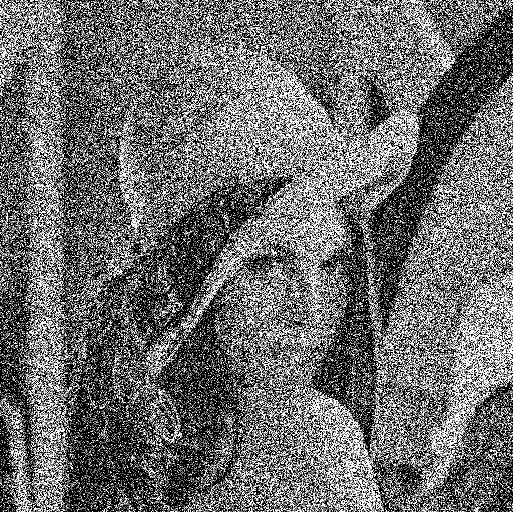}
}
\subfigure[{\scriptsize Model \eqref{overallModel} (24.28/0.687)}]{
\centering
\includegraphics[width=0.27\textwidth]{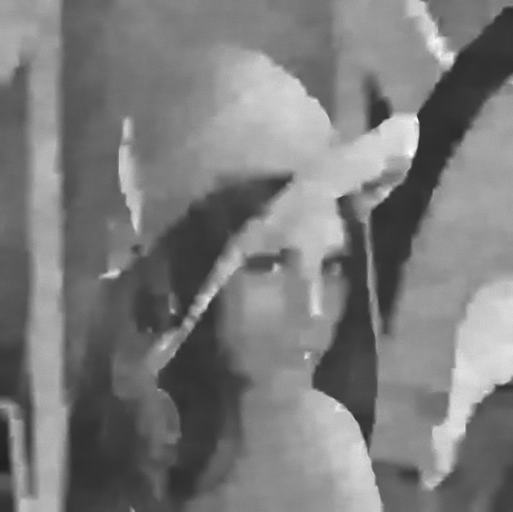}
}
\subfigure[{\tiny Model \eqref{lowcase} (\textbf{24.68}/\textbf{0.710})}]{
\centering
\includegraphics[width=0.27\textwidth]{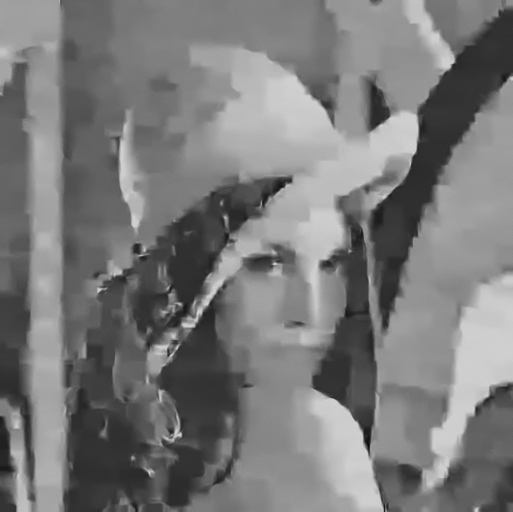}
}\\
\subfigure[{\scriptsize Noisy image (peak = 7)}]{
\centering
\includegraphics[width=0.27\textwidth]{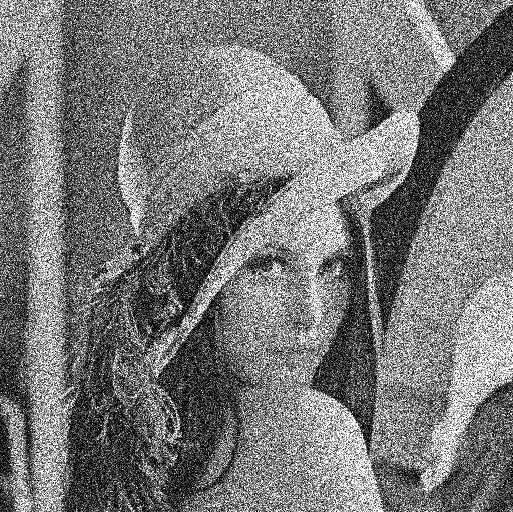}
}
\subfigure[{\tiny Model \eqref{overallModel} (\textbf{27.21}/\textbf{0.770})}]{
\centering
\includegraphics[width=0.27\textwidth]{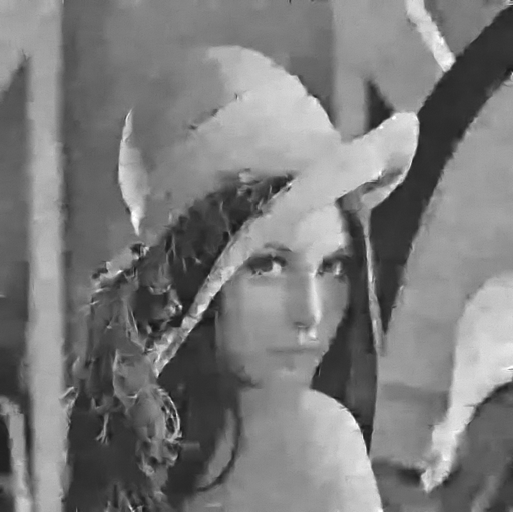}
}
\subfigure[{\scriptsize Model \eqref{lowcase} (27.00/0.760)}]{
\centering
\includegraphics[width=0.27\textwidth]{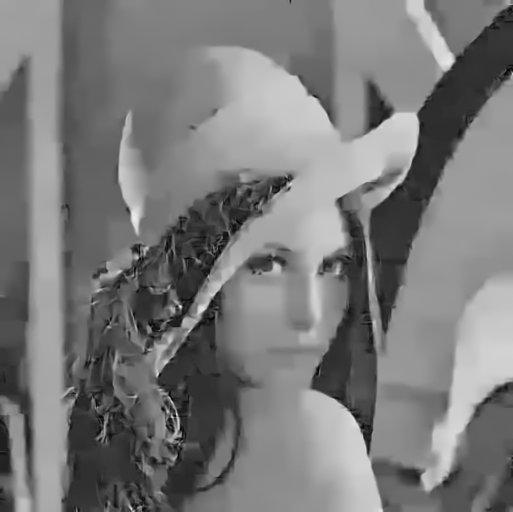}
}
\caption{\color{black}
Poisson denoising results with an ad hoc data term \eqref{lowcase}
and the Gaussian noise related data term \eqref{overallModel}.
The test image is corrupted by different peak values. As usual, in high-peak
Poisson denoising, the widely used Gaussian noise related data term
\eqref{overallModel} performs better; however, for the cases of low-peak (peak $< 5$), the
ad hoc data term \eqref{lowcase} is preferred.
The results are reported by PSNR/MSSIM index.}
\label{datatermcompare}
\end{figure*}

Even though the proposed model \eqref{overallModel} works quite well in cases
with relatively high peak value, in experiments, we find that its performance
for cases with low peak value (e.g., $\text{peak} < 5$)
is not satisfactory, as it generally leads to somehow over-smoothing result.
An example for the case of $\text{peak} = 2$ is shown in Figure
\ref{datatermcompare}(b). \textit{We believe that the reason is
attributed to the data term.
Note that the noise statistics of the transformed data can only be approximately treated as additive Gaussian noise.
The approximation error will dramatically increase for cases of low count.
Therefore, it is better to exploit a different data term for these cases, rather than employing the same quadratic data term.}

In practice, we find that a naive attempt to exploit the direct Poisson noise statistics
derived data term \eqref{poissonfidelity} can generally provide preferable results for these cases,
\ie, we replace the second step in the framework \eqref{overallModel} with the following variational model
\begin{equation}\label{lowcase}
\color{black}
\hat{u} =\arg\min\limits_{u} \suml{i=1}{N_f}\suml{p=1}{N} \rho((k_i * u)_p) + \lambda \langle u-v \mathrm{log}u,1 \rangle \,.
\end{equation}
Note that the FoE prior model is $\textbf{FoE}_{\textbf{A}}$, and this model
works in the Anscombe transform domain only for the cases of low-peak.

The result of this modified version is shown in Figure
\ref{datatermcompare}(c). Comparing the quantitative
indexes and the visual quality, we can see that when the peak value is very low,
the modified model \eqref{lowcase} generates
preferable result on detail preservation, especially around the tassel of $Lena$'s hat.

In order to further investigate the influence of the data term, we also consider a Poisson denoising experiment for the
case with relative high peak value, e.g., $\text{peak} = 7$ by using two different data terms, i.e., \eqref{overallModel} and
\eqref{lowcase}. The corresponding results are shown in Figure
\ref{datatermcompare}(e) and (f). One can see that for this case, the conventional
Gaussian noise derived data term performs better. As a consequence, it is reasonable
to adopt adaptive data terms in the proposed framework to accommodate
different situations.

In order to better validate this adaptation, we repeated the above
evaluation based on the
eight test images in Fig~\ref{testimages}, and summarize the average performance in terms of PSNR and MSSIM
in Table~\ref{datatermcomparefws}. Numerical results strength the adaptation described above.

\begin{figure}[t!]
\begin{center}
    {\includegraphics[width=0.5\textwidth]{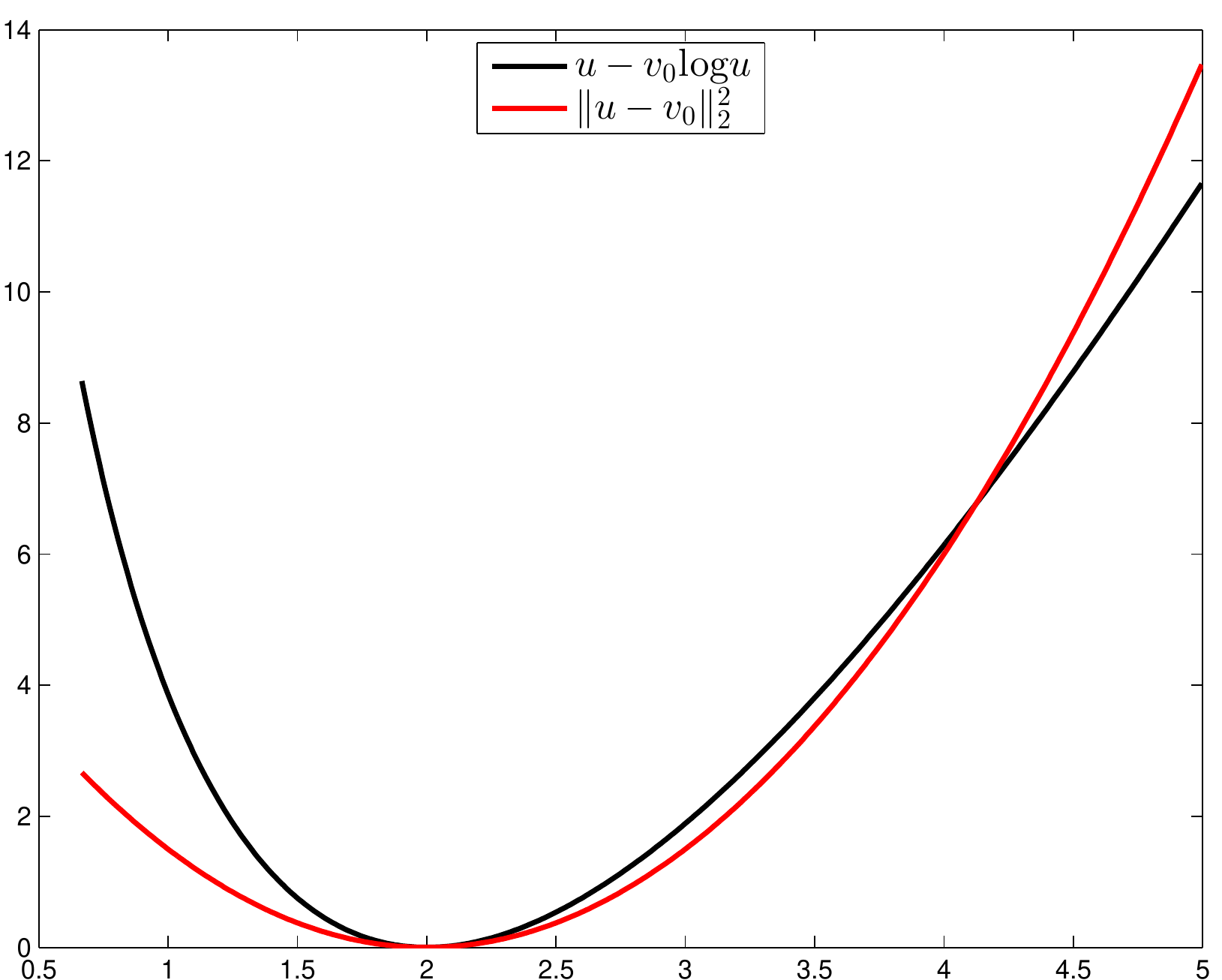}}
\end{center}
    \caption{A comparison of two penalty functions at a point $v_0 = 2$:
quadratic function and the I-divergence term in model \eqref{lowcase}.}
\label{fig:div}
\end{figure}

Although there is no clear interpretation why it is a good choice to replace the quadratic data term with
\eqref{lowcase} for the case of low peak, we can find some clues in Figure \ref{fig:div} for an explanation.
We can see that the I-divergence data term \eqref{lowcase} is quite similar to the quadratic data term at
the right part of the point $v_0$, while there is some difference at the left part, \ie, this modification makes a difference,
but not a huge difference.

{\color{black}
An additional Poisson denoising example for the case of $\text{peak} = 4$ using the model \eqref{lowcase} is shown
Figure \ref{lowpeak}(c). One can see that the resulting performance is even better
than the BM3D-based algorithm}
\cite{INAnscombe5}.

\begin{table}[t!]
\centering
\begin{tabular}{|c |c |c |}
\cline{1-3}
&\multicolumn{2}{c|}{PSNR/MSSIM}\\
\cline{2-3}
&\multicolumn{1}{c|}{Model \eqref{overallModel}} &\multicolumn{1}{c|}{Model \eqref{lowcase}}\\
\cline{1-3}
\multirow{1}*{peak=2}
&22.93/0.61 & \textbf{23.34}/\textbf{0.64}\\
\hline
\multirow{1}*{peak=7}
&\textbf{25.80}/\textbf{0.72} & 25.57/0.70\\
\cline{1-3}
\end{tabular}
\caption{\color{black}
The average recovery error in terms of PSNR (in dB) and MSSIM on the eight test images in Fig~\ref{testimages} by using two different data term. It is true that
for low-peak Poisson denoising, the ad hoc data term \eqref{lowcase} is preferred.}
\label{datatermcomparefws}
\vspace*{-0.35cm}
\end{table}

{\color{black}
\subsection{The final selected FoE-based variational model for Poisson denoising}
}After a comprehensive investigation of several variants to incorporate the
FoE prior model for Poisson denoising,
we arrive at the following conclusions:\textit{
\begin{enumerate}
\item [A)] It is better to formulate the FoE-based Poisson denoising model in the Anscombe transform domain.
\item [B)] An adaptive modeling of the data term for different noise level is helpful.
\end{enumerate}}
As a rule of thumb, $\text{peak} = 5$ is used as the threshold. Just as indicated in Table \ref{datatermcomparefws}, for $\text{peak} \geq 5$, we employ the model \ref{overallModel}, and for $\text{peak} < 5$ we use the model \ref{lowcase}.
Therefore,
our best performing version is given as
\begin{equation}\label{FOEscheme}
 \hat{u} = \left\{
\begin{array}{ll}
\arg\min\limits_{u} E_{\textbf{FoE}_{\textbf{A}}}(u) + \frac \lambda 2 \|u - v\|^2_2 \,, &\text{peak} \geq 5\\
\arg\min\limits_{u} E_{\textbf{FoE}_{\textbf{A}}}(u) + \lambda \langle u-v \mathrm{log}u,1 \rangle \,, &
\text{peak} < 5
\end{array} \right.
\end{equation}
with $v = f_{Anscombe}\left( y \right)$ and $\hat x = \cI_C(\hat{u})$. 
{Note that, in the binning case, the threshold is still peak = 5. However, the peak value should be determinated by the binned image.} In the following test procedure, we will only present the results obtained by this final version for the sake of brevity.

\section{Comparison to State-of-the-arts}
\begin{figure*}[t!]
\centering
    \subfigure[{\scriptsize Boat ($512 \times 512$)}]{\includegraphics[width=0.23\textwidth]{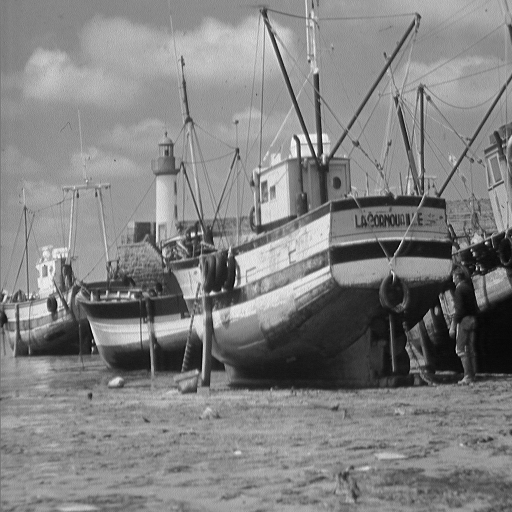}}
    \subfigure[{\scriptsize Cameraman ($256\times256$)}]{\includegraphics[width=0.23\textwidth]{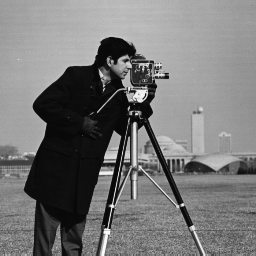}}
    \subfigure[{\scriptsize Fluocells ($512 \times 512$)}]{\includegraphics[width=0.23\textwidth]{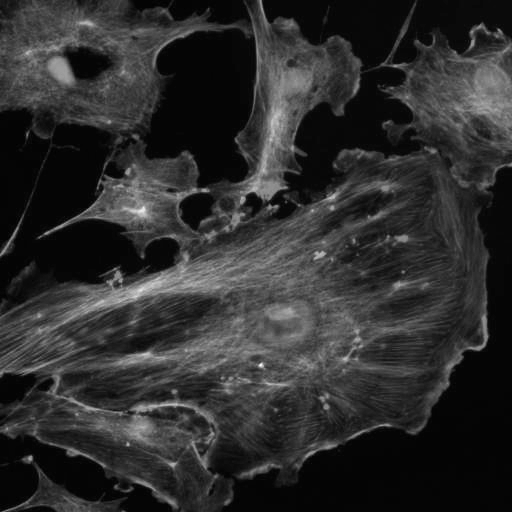}}
    \subfigure[{\scriptsize House ($256 \times 256$)}]{\includegraphics[width=0.23\textwidth]{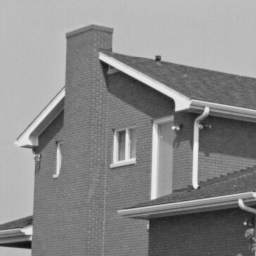}}\\
    \subfigure[{\scriptsize Lena ($512 \times 512$)}]{\includegraphics[width=0.23\textwidth]{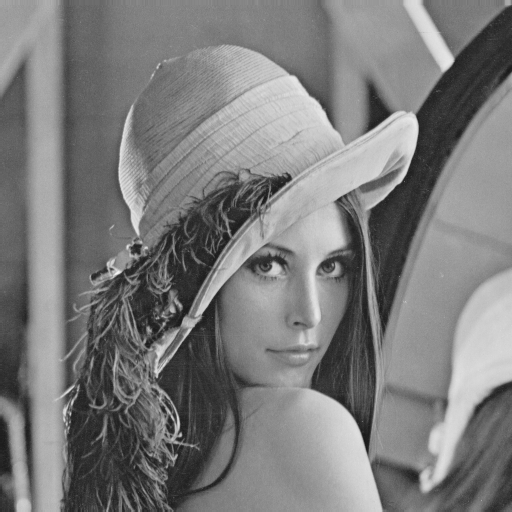}}
    \subfigure[{\scriptsize Bridge ($256 \times 256$)}]{\includegraphics[width=0.23\textwidth]{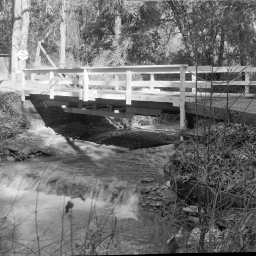}}
    \subfigure[{\scriptsize Pepper ($256 \times 256$)}]{\includegraphics[width=0.23\textwidth]{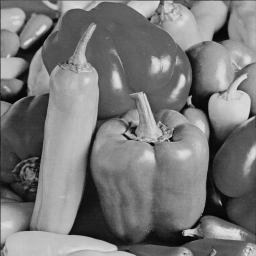}}
    \subfigure[{\scriptsize Man ($512 \times 512$)}]{\includegraphics[width=0.23\textwidth]{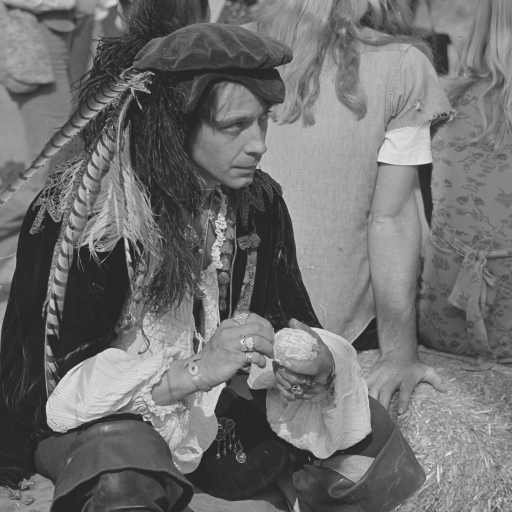}}\\
    \caption{Eight test images}\label{testimages}
\end{figure*}

In this section, we report numerical experiments on Poisson noise reduction with 8 test images in Fig.~\ref{testimages}.
Additionally, we compare
the proposed algorithm with the NLSPCA \cite{Salmon2} and the BM3D with the exact unbiased inverse Anscombe
\cite{INAnscombe5} (both with and without binning), which usually provide state-of-the-art results.
The corresponding codes are downloaded from the author's homepage, and we use them as is.
For the binning technique, we closely follow \cite{Salmon2} and
use a $3 \times 3$ ones kernel to increase the peak value to be 9 times higher, and
a bilinear interpolation for the upscaling of the low-resolution
recovered image.
Moreover, in our study we take one particular noise realization with the Matlab command
\textit{randn('seed',0);  rand('seed',0)}, because we want to provide determinant and easily reproducible results for
a direct comparison between different algorithms.
Two commonly used quality measures are taken to evaluate the Poisson denoising performance, \ie, PSNR and
the mean structural similarity index (MSSIM) \cite{SSIM}. For convenience sake, the proposed model \eqref{FOEscheme}
is abbreviated as FoEPNRbin if the binning technique is adopted, and FoEPNR otherwise.

\begin{figure*}[t!]
\centering
\subfigure[{\scriptsize $Cameraman$ image}]{
\centering
\includegraphics[width=0.22\textwidth]{figures/Fig.1-testimages/cameraman.png}
}
\subfigure[{\scriptsize NLSPCA (17.65/0.37)}]{
\centering
\includegraphics[width=0.22\textwidth]{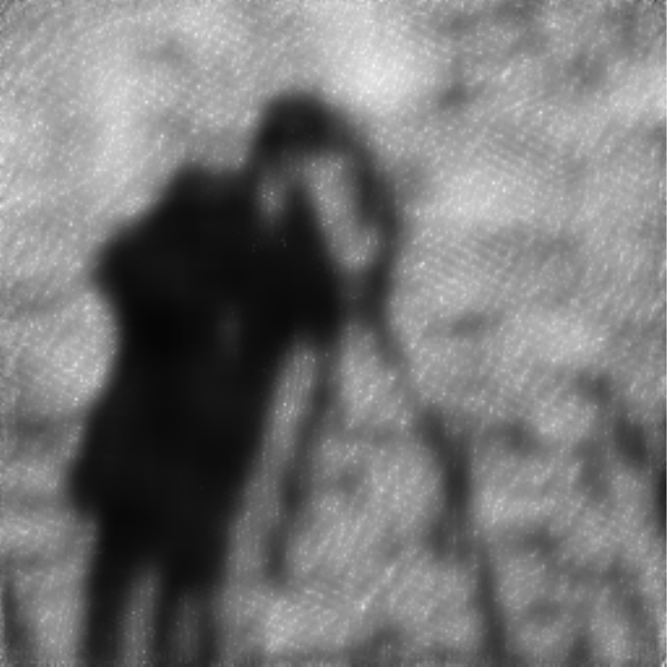}
}
\subfigure[{\scriptsize BM3D (17.06/0.42)}]{
\centering
\includegraphics[width=0.22\textwidth]{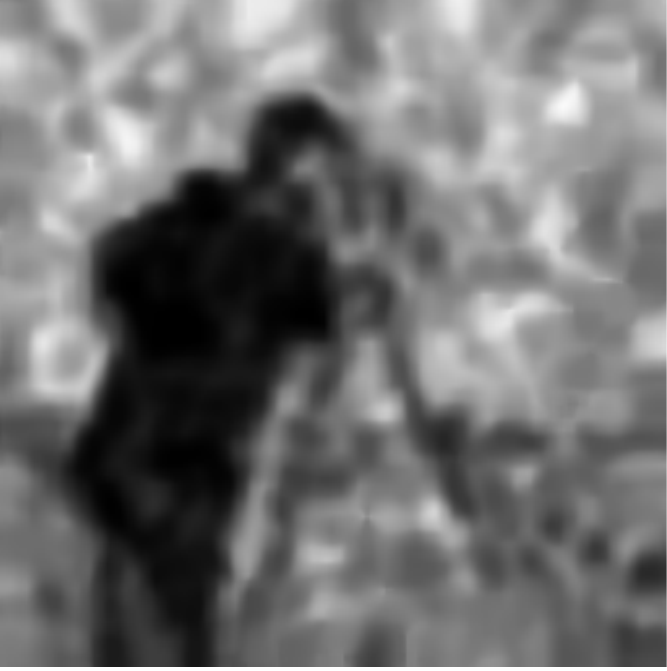}
}
\subfigure[{\scriptsize FoEPNR (17.85/0.49)}]{
\centering
\includegraphics[width=0.22\textwidth]{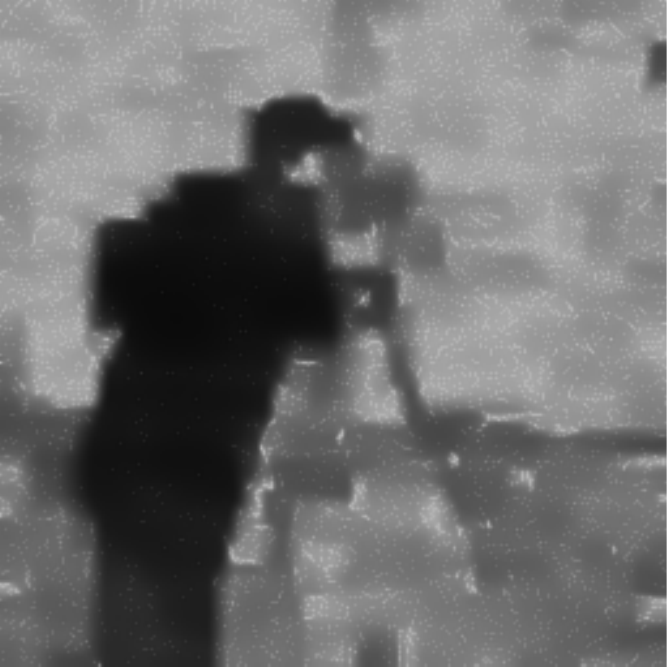}
}\\
\subfigure[{\scriptsize Noisy image. Peak=0.2}]{
\centering
\includegraphics[width=0.22\textwidth]{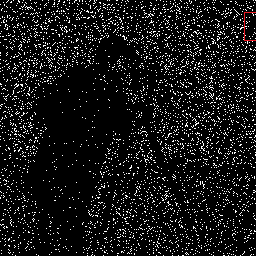}
}
\subfigure[{\scriptsize NLSPCAbin (17.82/0.53)}]{
\centering
\includegraphics[width=0.22\textwidth]{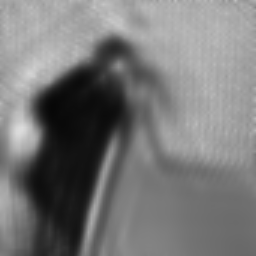}
}
\subfigure[{\scriptsize BM3Dbin (18.08/0.54)}]{
\centering
\includegraphics[width=0.22\textwidth]{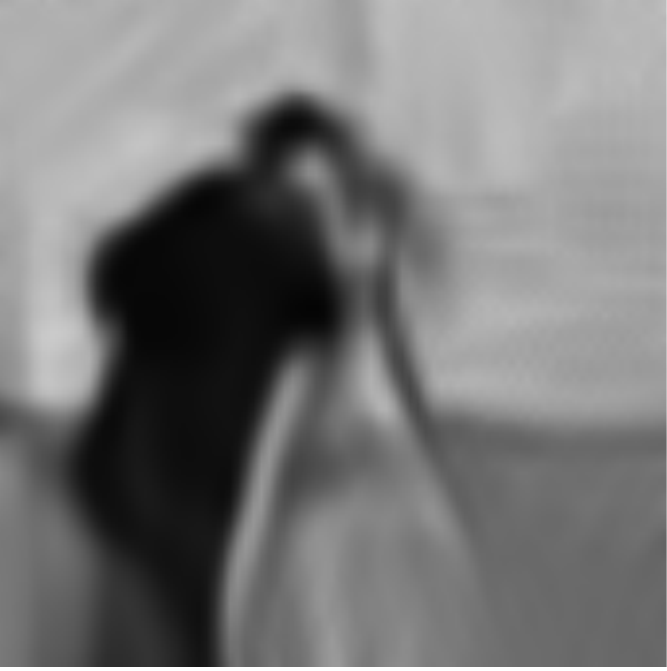}
}
\subfigure[{\scriptsize FoEPNRbin (\textbf{18.52}/\textbf{0.58})}]{
\centering
\includegraphics[width=0.22\textwidth]{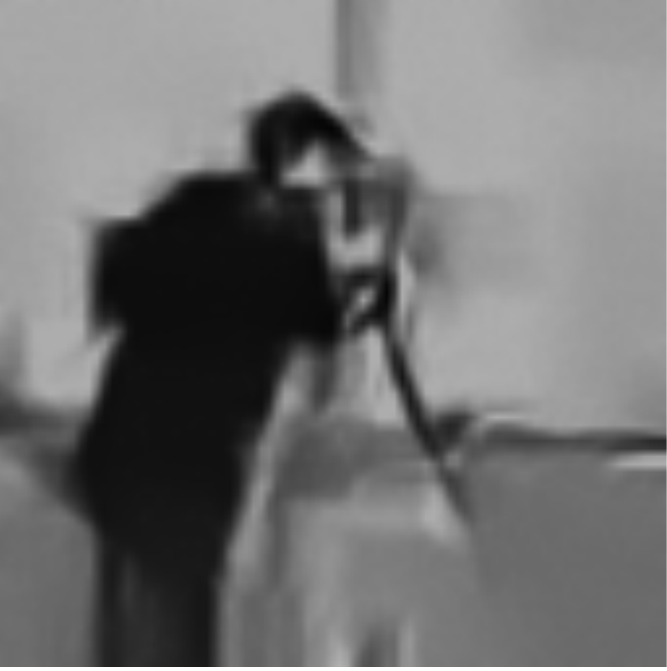}
}
\caption{Denoising of $cameraman$ image with peak=0.2. The PSNR is of the presented recovered images. Best results are marked.}
\label{peak0.2}
\end{figure*}

\begin{figure*}[t]
\centering
\subfigure[{\scriptsize $Pepper$ image}]{
\centering
\includegraphics[width=0.22\textwidth]{figures/Fig.1-testimages/peppers256used.png}
}
\subfigure[{\scriptsize NLSPCA (19.32/0.61)}]{
\centering
\includegraphics[width=0.22\textwidth]{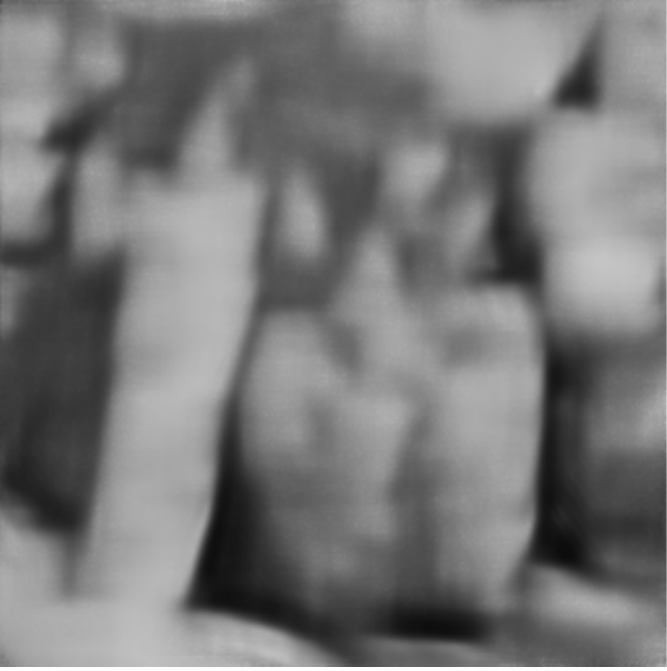}
}
\subfigure[{\scriptsize BM3D (20.00/0.62)}]{
\centering
\includegraphics[width=0.22\textwidth]{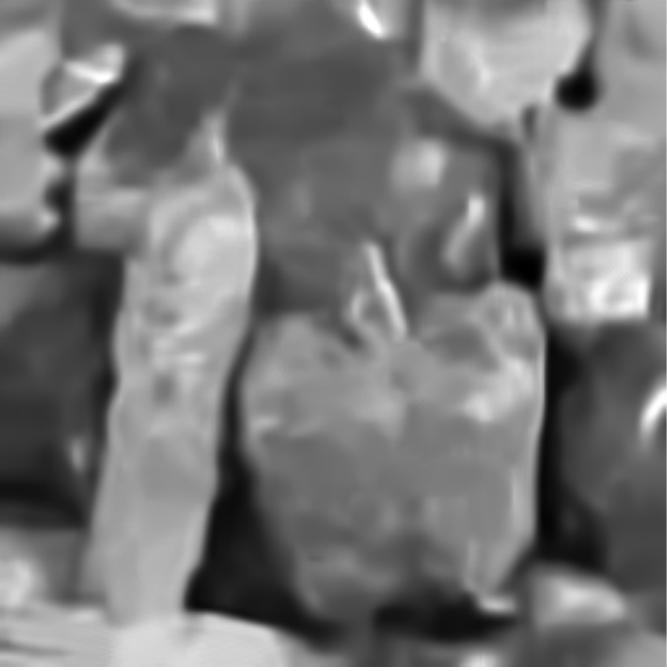}
}
\subfigure[{\scriptsize FoEPNR (\textbf{20.39}/\textbf{0.66})}]{
\centering
\includegraphics[width=0.22\textwidth]{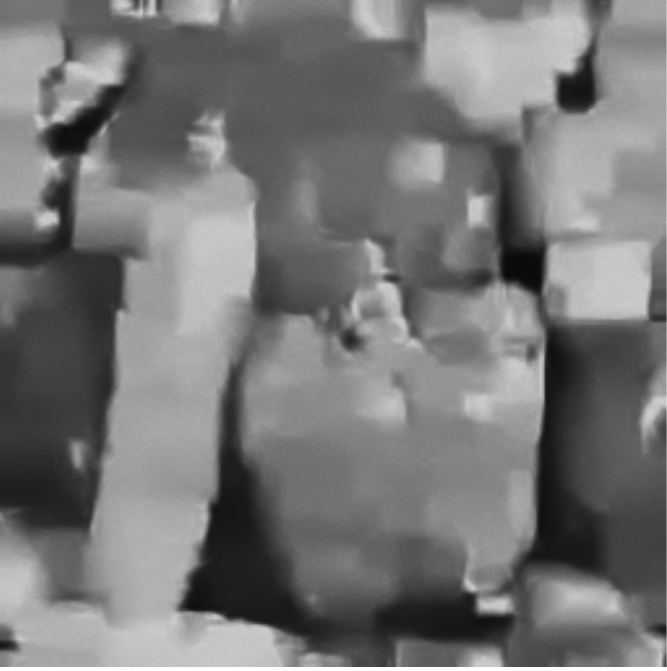}
}\\
\subfigure[{\scriptsize Noisy image. Peak=1}]{
\centering
\includegraphics[width=0.22\textwidth]{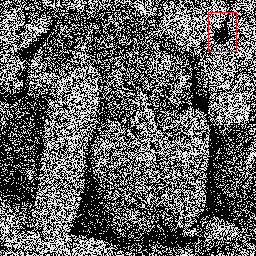}
}
\subfigure[{\scriptsize NLSPCAbin (17.01/0.54)}]{
\centering
\includegraphics[width=0.22\textwidth]{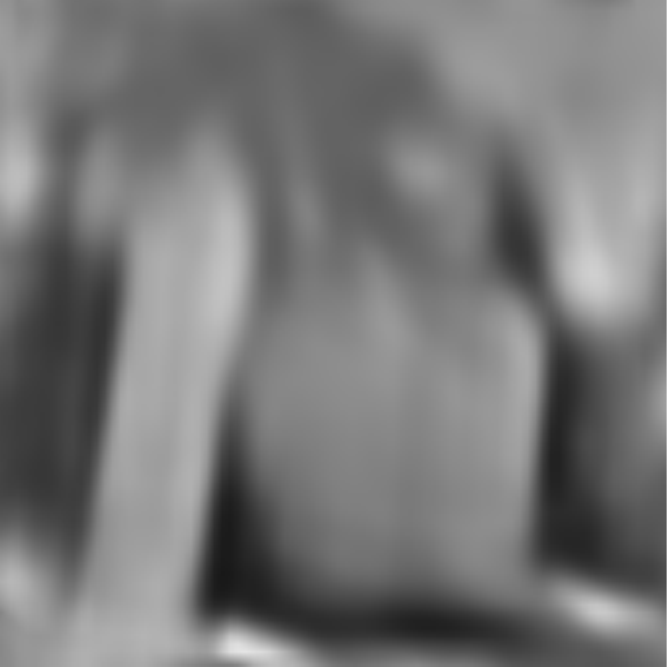}
}
\subfigure[{\scriptsize BM3Dbin (20.14/0.64)}]{
\centering
\includegraphics[width=0.22\textwidth]{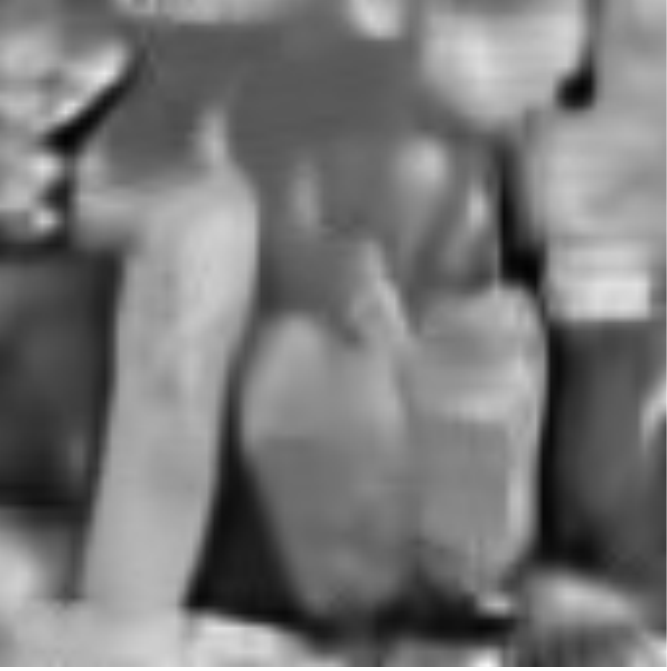}
}
\subfigure[{\scriptsize FoEPNRbin (19.92/0.64)}]{
\centering
\includegraphics[width=0.22\textwidth]{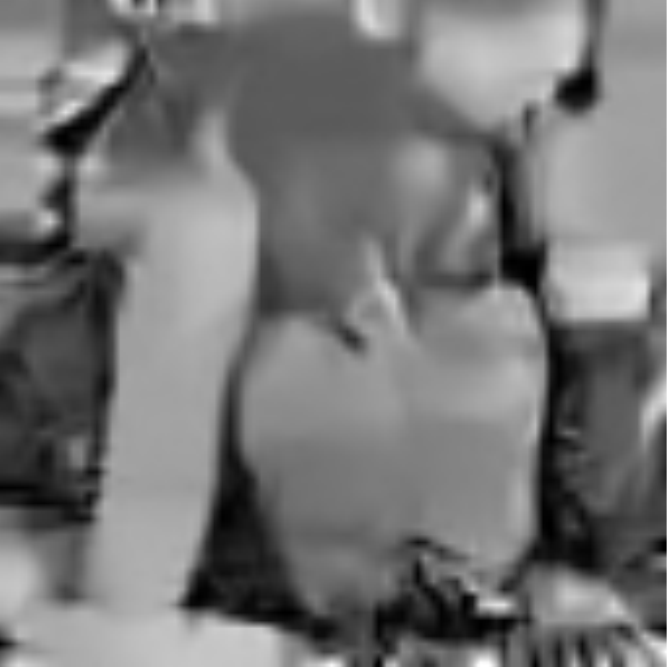}
}
\caption{Denoising of $pepper$ image with peak=1. The results are reported by PSNR/MSSIM index. Best results are marked.}
\label{peak1}
\end{figure*}

\subsection{Experimental results}
We evaluated the performance of the proposed variational model \eqref{FOEscheme} (with and without binning) for various images in Fig.~\ref{testimages} with different peak values ranging from 0.1 to 40.

Examining the recovery images in Fig.~\ref{peak0.2}, we see that compared with the other methods, the proposed algorithm FoEPNRbin is more accurate at capturing strong edges and details, especially the bright part. It is worthy noting that, by closely observation, for the low peak value the proposed method with binning technique performs better on capturing the structure of the image, and therefore provides higher PSNR/MSSIM value.

\begin{table*}[h]
\tiny
\centering
\begin{tabular}{|p{1.2cm}|p{0.35cm}|p{1.1cm}|p{1.1cm}|p{1.1cm}|p{1.1cm}|p{1.1cm}|p{1.1cm}|p{1.1cm}|p{1.1cm}|p{1.1cm}|}
\hline
Method & Peak & Boat & Cameraman & Fluocells & House & Lena & Bridge & Pepper & Man &Average value\\
\hline
\hline
\multirow{6}{1.2cm}{NLSPCA\\NLSPCAbin\\BM3D\\BM3Dbin\\FoEPNR\\FoEPNRbin}
& &18.27/0.41&16.56/0.40&21.42/0.49&17.44/0.44&18.30/0.47&16.60/0.22&15.62/0.39&18.15/0.41&17.80/0.40
\\
& &18.80/0.45&17.19/0.48&21.45/0.50&18.09/0.52&19.03/\textbf{0.59}&16.92/0.22&15.19/0.45&18.41/0.46&18.14/0.46
\\
&0.1&16.34/0.33&15.65/0.34&20.68/0.42&15.92/0.38&16.78/0.42&15.96/0.21&15.22/0.42&16.71/0.36&16.66/0.36
\\
& &18.89/0.46&16.99/\textbf{0.51}&21.45/0.51&17.84/0.57&18.78/\textbf{0.59}&17.03/0.22&15.44/0.49&18.31/\textbf{0.47}&18.09/0.48
\\
& &18.68/0.42&16.81/0.41&21.41/0.48&18.23/0.50&18.26/0.49&\textbf{17.35}/0.22&15.61/0.43&18.37/0.41&18.09/0.42
\\
& &\textbf{19.28}/\textbf{0.47}&\textbf{17.66}/\textbf{0.51}&\textbf{21.70}/\textbf{0.52}&\textbf{18.47}/\textbf{0.61}&
\textbf{19.07}/\textbf{0.59}&17.25/\textbf{0.23}&\textbf{16.07}/\textbf{0.51}&\textbf{18.68}/\textbf{0.47}&\textbf{18.52}/\textbf{0.49}
\\
\hline
\multirow{6}{1.2cm}{NLSPCA\\NLSPCAbin\\BM3D\\BM3Dbin\\FoEPNR\\FoEPNRbin}
& &19.11/0.41&17.65/0.37&22.45/0.53&18.44/0.42&19.54/0.50&17.45/0.24&16.94/0.44&19.18/0.43&18.85/0.42
\\
& &19.58/0.47&17.82/0.53&21.98/0.52&\textbf{19.15}/0.57&19.73/0.61&17.45/0.24&16.24/0.51&19.01/0.48&18.87/0.49
\\
&0.2&18.38/0.40&17.06/0.42&22.23/0.51&17.89/0.48&18.96/0.50&17.09/0.25&16.67/0.50&18.68/0.44&18.40/0.46
\\
& &19.99/0.47&18.08/0.54&22.46/0.55&19.03/0.58&20.29/0.61&\textbf{17.82}/0.25&16.73/0.52&19.66/0.68&19.26/\textbf{0.53}
\\
& &19.78/0.43&17.85/0.49&22.19/0.54&19.03/0.56&19.35/0.56&17.81/0.25&16.36/0.50&19.39/0.47& 18.97/0.48
\\
& &\textbf{20.03}/\textbf{0.49}&\textbf{18.52}/\textbf{0.58}&\textbf{22.58}/\textbf{0.57}&18.54/\textbf{0.63}&
\textbf{20.53}/\textbf{0.64}&17.53/\textbf{0.26}&\textbf{17.07}/\textbf{0.54}&\textbf{19.76}/\textbf{0.50}& \textbf{19.32}/\textbf{0.53}
\\
\hline
\multirow{6}{1.2cm}{NLSPCA\\NLSPCAbin\\BM3D\\BM3Dbin\\FoEPNR\\FoEPNRbin}
& &20.42/0.45&19.06/0.51&23.73/0.59&20.29/0.54&21.26/0.57&18.46/0.28&18.34/0.55&20.60/0.49&20.27/0.50
\\
& &20.17/0.48&18.31/0.54&22.51/0.54&20.47/0.61&20.57/0.62&18.19/0.26&16.93/0.53&19.74/0.49&19.61/0.51
\\
&0.5&20.07/0.47&18.64/0.52&23.81/0.58&19.71/0.58&21.13/0.59&18.15/0.28&18.18/0.56&20.40/0.49&20.07/\textbf{0.57}
\\
& &21.06/0.50&19.53/0.58&23.85/0.60&\textbf{21.25}/0.64&\textbf{22.37}/0.64&18.53/0.28&18.55/0.59&\textbf{21.09}/\textbf{0.52}
&20.78/0.54
\\
& &21.01/0.50&19.29/0.57&23.57/0.58&20.95/0.64&21.95/0.63&\textbf{18.59}/\textbf{0.29}&\textbf{18.71}/0.58&20.96/0.51&20.63/0.54
\\
& &\textbf{21.11}/\textbf{0.51}&\textbf{19.85}/\textbf{0.62}&\textbf{24.03}/\textbf{0.63}
&21.18/\textbf{0.66}&22.21/\textbf{0.66}&18.49/\textbf{0.29}&18.66/\textbf{0.60}&21.01/\textbf{0.52}&\textbf{20.82}/0.56
\\
\hline

\multirow{6}{1.2cm}{NLSPCA\\NLSPCAbin\\BM3D\\BM3Dbin\\FoEPNR\\FoEPNRbin}
& &21.17/0.49&20.26/0.61&24.61/0.63&21.54/0.61&22.53/0.64&18.98/0.29&19.32/0.61&21.41/0.52&21.23/0.55
\\
& &20.20/0.48&18.41/0.55&22.64/0.54&20.64/0.63&20.66/0.62&18.30/0.26&17.01/0.54&19.88/0.49&19.72/0.51
\\
&1&21.40/0.51&20.40/0.60&24.65/0.62&21.80/0.65&22.65/0.64&19.38/0.33&20.00/0.62&21.64/0.53&21.49/0.56
\\
& &21.90/0.52&20.49/0.62&24.86/0.64&\textbf{22.68}/\textbf{0.69}&\textbf{23.54}/0.68&
19.53/0.32&20.14/0.64&22.00/0.54& 21.89/0.58
\\
& &\textbf{22.01}/\textbf{0.53}&\textbf{20.85}/0.63&24.87/0.64&22.41/0.68&23.28/0.67&\textbf{19.73}/\textbf{0.34}&
\textbf{20.39}/\textbf{0.66}&\textbf{22.04}/\textbf{0.55}&\textbf{21.95}/\textbf{0.59}
\\
& &21.88/0.52&20.73/\textbf{0.64}&\textbf{25.09}/\textbf{0.65}&22.18/0.68&23.35/\textbf{0.69}&19.32/0.33&
19.92/0.64&21.84/0.54&21.79/\textbf{0.59}
\\
\hline

\multirow{6}{1.2cm}{NLSPCA\\NLSPCAbin\\BM3D\\BM3Dbin\\FoEPNR\\FoEPNRbin}
& &21.76/0.52&20.69/0.63&25.34/0.65&23.23/0.70&23.88/0.68&19.49/0.31&20.25/0.65&22.33/0.55&22.12/0.59
\\
& &20.29/0.48&18.33/0.55&22.69/0.54&20.85/0.64&20.91/0.63&18.32/0.26&17.12/0.54&19.88/0.49& 19.80/0.52
\\
&2&22.70/0.55&22.07/0.64&26.06/0.68&23.38/0.66&24.16/0.66&20.25/\textbf{0.39}&21.73/0.67&23.01/0.58&22.92/0.60
\\
& &22.78/0.56&21.34/0.66&26.00/0.67&23.90/0.71&\textbf{24.69}/0.71&20.15/0.35&21.35/0.68&22.98/0.58&22.90/0.62
\\
& &\textbf{23.15}/\textbf{0.58}&\textbf{22.74}/\textbf{0.71}&\textbf{26.25}/\textbf{0.69}&
\textbf{24.09}/\textbf{0.73}&24.68/0.71&\textbf{20.51}/\textbf{0.39}&\textbf{22.05}/\textbf{0.72}&
\textbf{23.27}/\textbf{0.60}&\textbf{23.34}/\textbf{0.64}
\\
& &22.76/0.55&21.51/0.64&26.07/0.67&23.55/0.70&24.59/\textbf{0.72}&20.04/0.38&21.43/0.68&22.90/0.56&22.86/0.61
\\
\hline

\multirow{6}{1.2cm}{NLSPCA\\NLSPCAbin\\BM3D\\BM3Dbin\\FoEPNR\\FoEPNRbin}
& &22.43/0.54&21.04/0.65&26.33/0.68&24.28/0.72&24.48/0.70&20.19/0.35&20.95/0.67&22.91/0.57&22.83/0.61
\\
& &20.29/0.48&18.33/0.55&22.68/0.54&20.92/0.64&20.78/0.63&18.37/0.26&16.92/0.54&19.86/0.49&19.77/0.52
\\
&4&24.24/0.61&23.96/0.71&\textbf{27.63}/\textbf{0.73}&25.73/0.72&26.05/0.71&\textbf{21.54}/\textbf{0.47}&
\textbf{23.62}/0.74&\textbf{24.37}/\textbf{0.63}& 24.64/0.66
\\
& &23.66/0.59&21.95/0.68&26.96/0.70&25.08/0.74&25.77/0.74&20.80/0.39&22.40/0.72&23.84/0.61&23.81/0.65
\\
& &\textbf{24.33}/\textbf{0.62}&\textbf{24.41}/\textbf{0.76}&27.55/\textbf{0.73}&\textbf{26.03}/\textbf{0.77}&
\textbf{26.11}/\textbf{0.75}&21.51/0.45&23.55/\textbf{0.77}&24.31/\textbf{0.63}&\textbf{24.73}/\textbf{0.69}
\\
& &23.61/0.58&22.08/0.66&27.07/0.68&25.08/0.74&25.66/0.74&20.92/0.42&22.57/0.72&23.87/0.60&23.86/0.64
\\
\hline
\multirow{6}{1.2cm}{NLSPCA\\NLSPCAbin\\BM3D\\BM3Dbin\\FoEPNR\\FoEPNRbin}
& & 22.94/0.56 & 21.16/0.65 & 26.99/0.69& 24.49/0.73& 25.10/0.72& 20.36/0.35&20.92/0.67 &23.32/0.58 & 23.16/0.62
\\
& &20.02/0.47& 18.19/0.55 & 22.49/0.53 &20.17/0.63& 19.13/0.60&17.78/0.24&15.97/0.51&19.49/0.48&19.16/0.50
\\
&40&\textbf{29.39}/\textbf{0.79}&\textbf{29.17}/\textbf{0.85} &\textbf{32.62}/\textbf{0.88}&\textbf{32.06}/\textbf{0.86}&\textbf{31.68}/\textbf{0.86}& 25.58/0.73&\textbf{29.64}/\textbf{0.88}& \textbf{28.95}/\textbf{0.80}&\textbf{29.89}/\textbf{0.83}
\\
& &25.56/0.68&23.07/0.73& 29.93/0.81&27.36/0.79&28.44/0.81&22.63/0.54&24.87/0.81&26.13/0.71&26.00/0.74
\\
& &29.20/0.78&28.93/\textbf{0.85}&32.43/0.87 & 31.31/0.85&31.33/0.85&\textbf{25.79}/\textbf{0.76}&29.55/\textbf{0.88}
&28.92/\textbf{0.80}&29.68/0.80
\\
& &25.63/0.69&22.92/0.74&30.08/0.82&27.43/0.79&28.45/0.81&22.73/0.56&24.95/0.82&26.25/0.72&26.06/0.74
\\
\hline
\end{tabular}
\caption{Comparison of the performance of the test algorithms in terms of PSNR and MSSIM. Best results are marked.}
\label{resultshow}
\end{table*}

\begin{figure*}[t]
\centering
\subfigure[{\scriptsize $Boat$ image}]{
\centering
\includegraphics[width=0.22\textwidth]{figures/Fig.1-testimages/boat.png}
}
\subfigure[{\scriptsize NLSPCA (21.76/0.52)}]{
\centering
\includegraphics[width=0.22\textwidth]{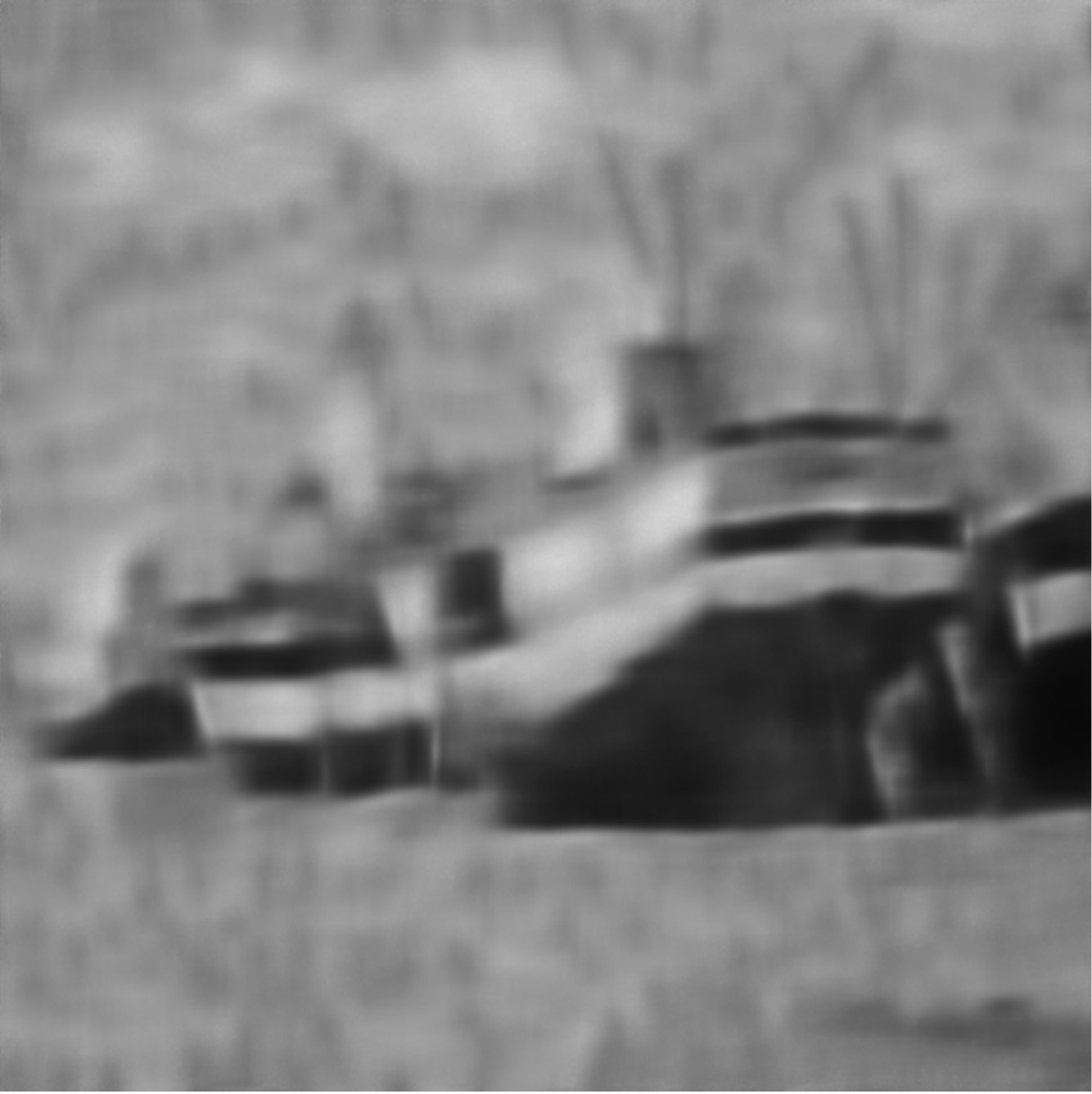}
}
\subfigure[{\scriptsize BM3D (22.70/0.55)}]{
\centering
\includegraphics[width=0.22\textwidth]{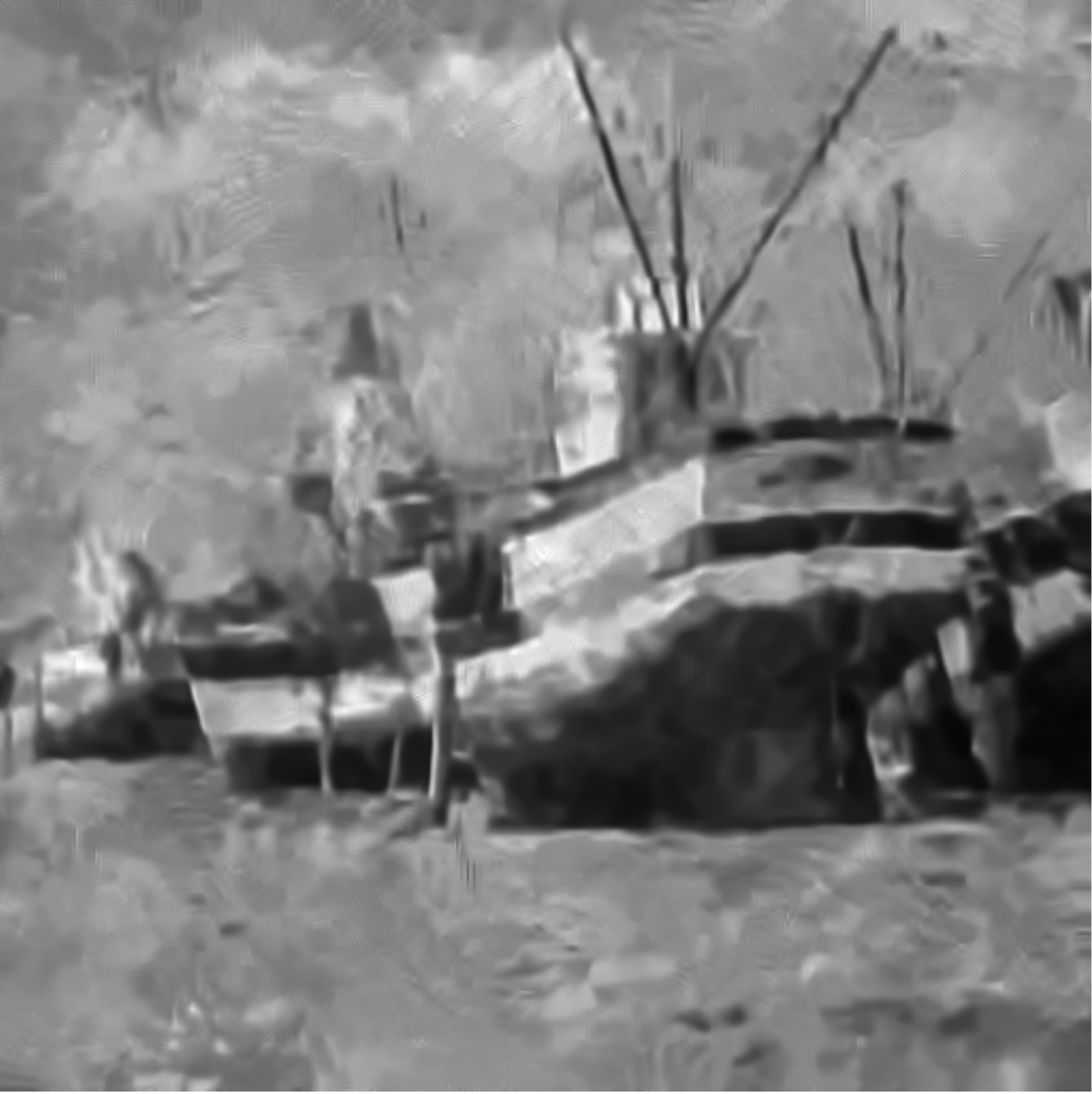}
}
\subfigure[{\scriptsize FoEPNR (\textbf{23.15}/\textbf{0.58})}]{
\centering
\includegraphics[width=0.22\textwidth]{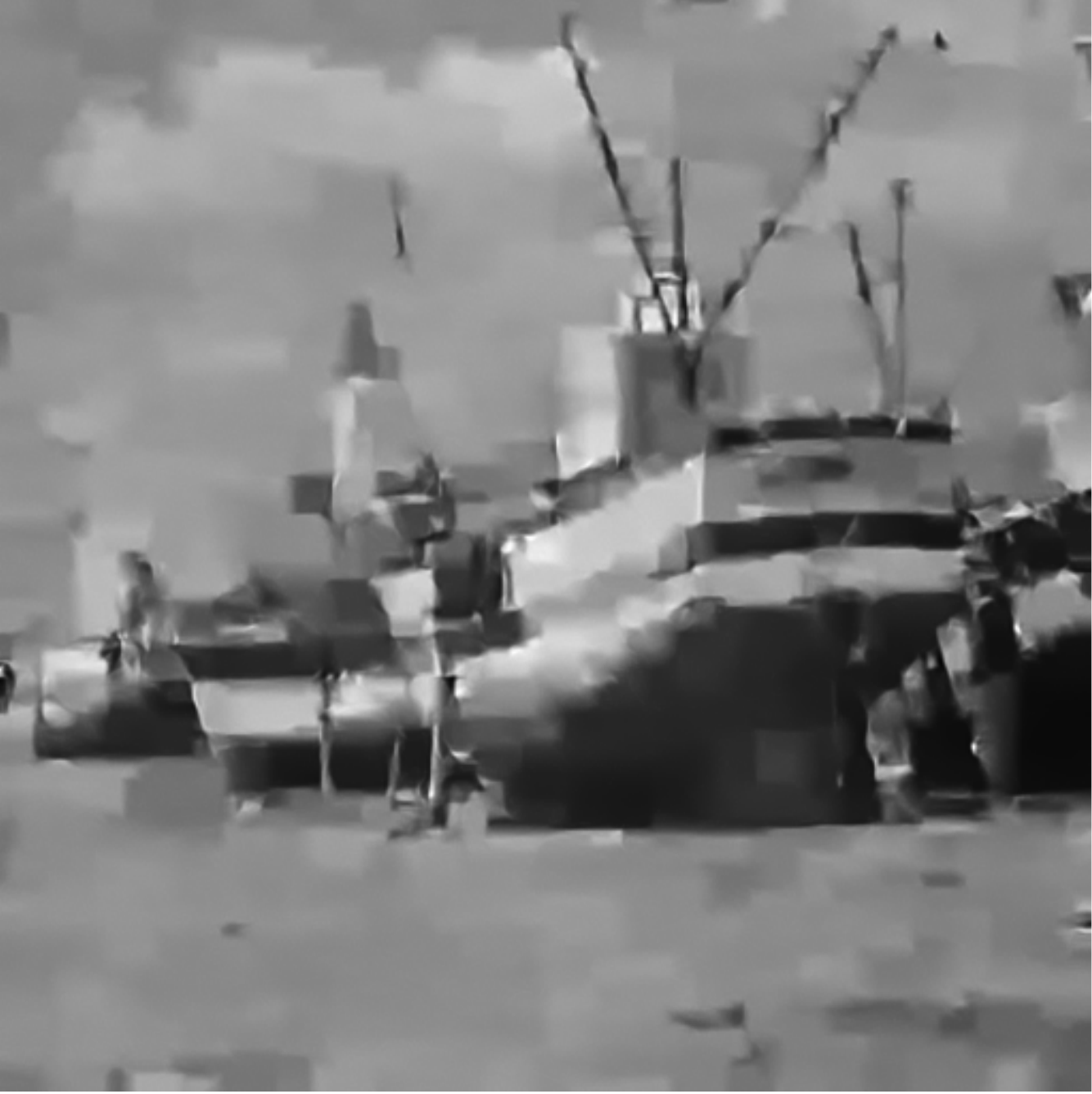}
}\\
\subfigure[{\scriptsize Noisy image. Peak=2}]{
\centering
\includegraphics[width=0.22\textwidth]{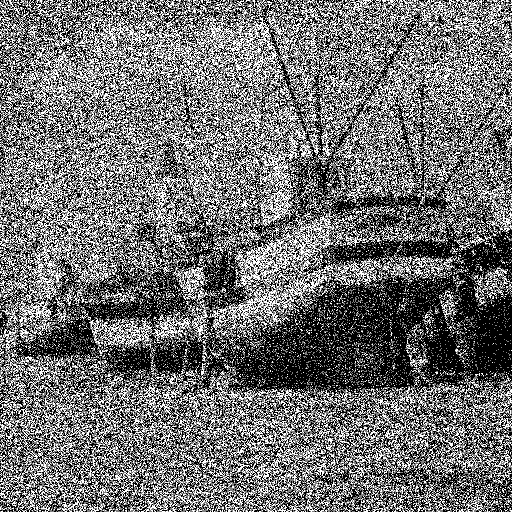}
}
\subfigure[{\scriptsize NLSPCAbin (20.29/0.48)}]{
\centering
\includegraphics[width=0.22\textwidth]{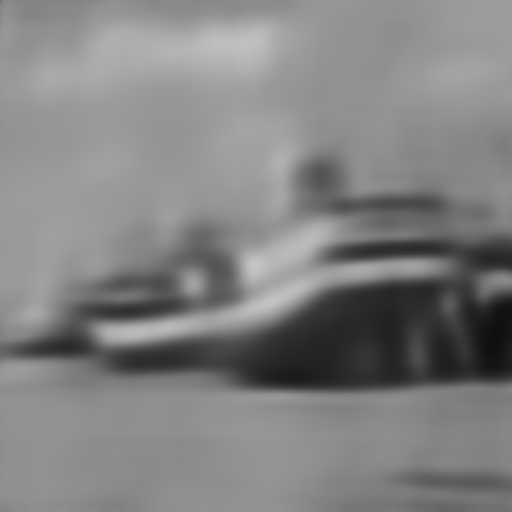}
}
\subfigure[{\scriptsize BM3Dbin (22.78/0.56)}]{
\centering
\includegraphics[width=0.22\textwidth]{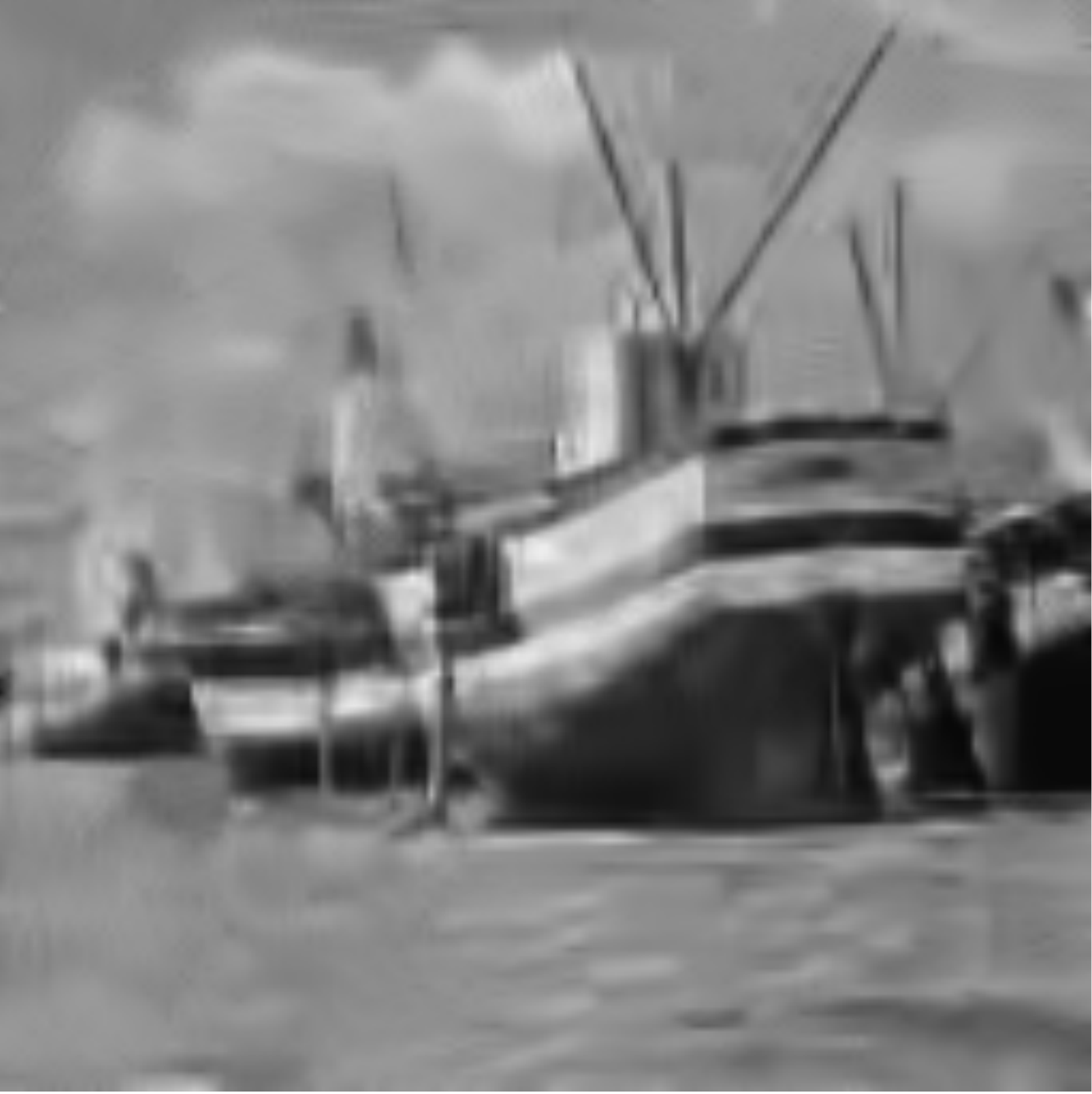}
}
\subfigure[{\scriptsize FoEPNRbin (22.76/0.55)}]{
\centering
\includegraphics[width=0.22\textwidth]{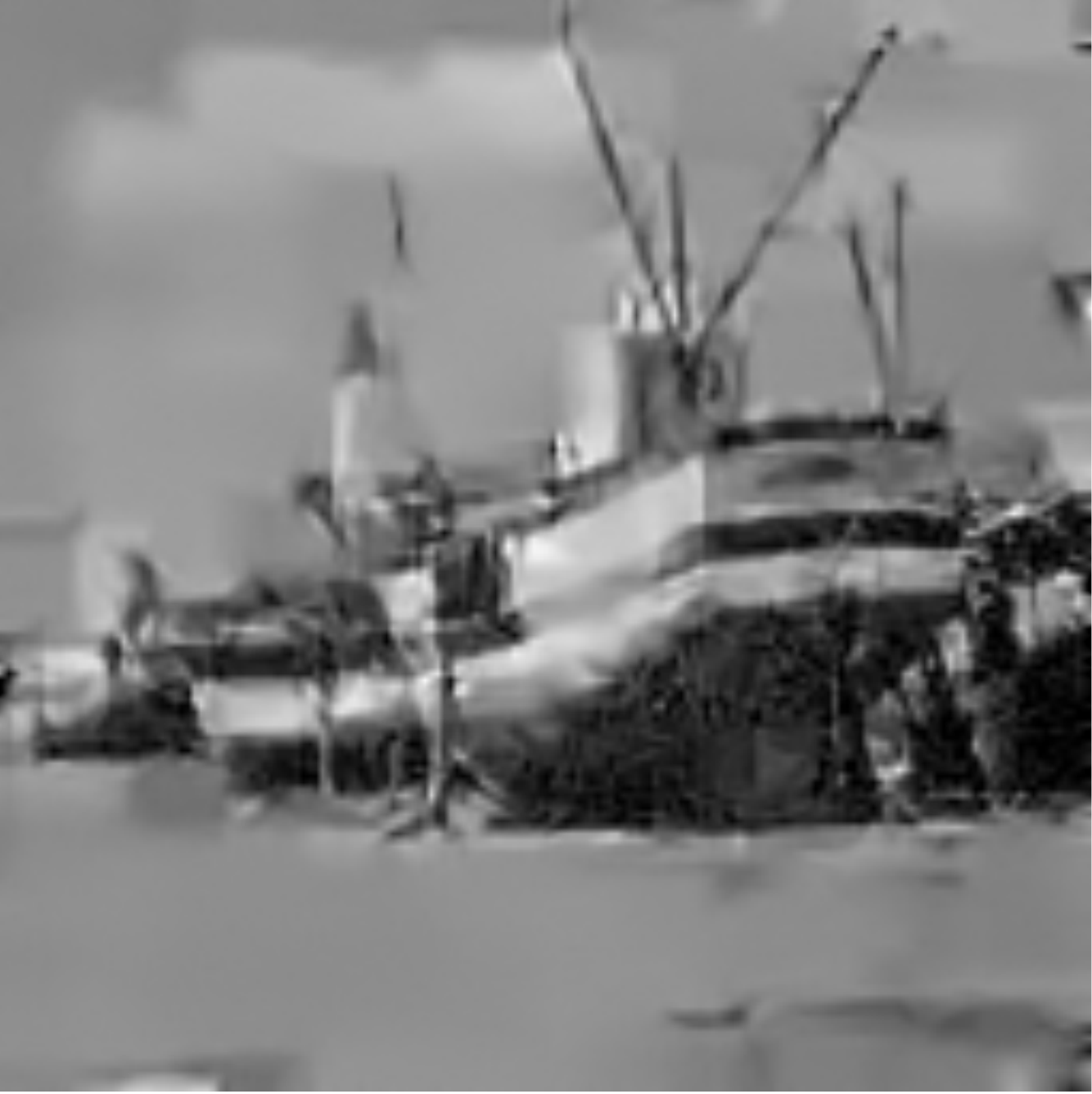}
}
\caption{Denoising of $boat$ image with peak=2. The results are reported by PSNR/MSSIM index. Best results are marked.}
\label{peak2}
\end{figure*}

\begin{figure*}[t]
\centering
\subfigure[{\scriptsize $House$ image}]{
\centering
\includegraphics[width=0.22\textwidth]{figures/Fig.1-testimages/house.png}
}
\subfigure[{\scriptsize NLSPCA (24.28/0.72)}]{
\centering
\includegraphics[width=0.22\textwidth]{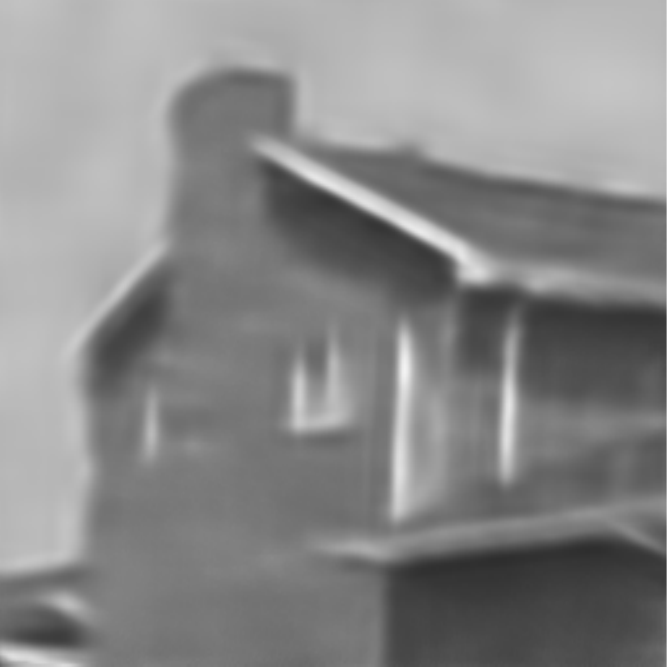}
}
\subfigure[{\scriptsize BM3D (25.73/0.72)}]{
\centering
\includegraphics[width=0.22\textwidth]{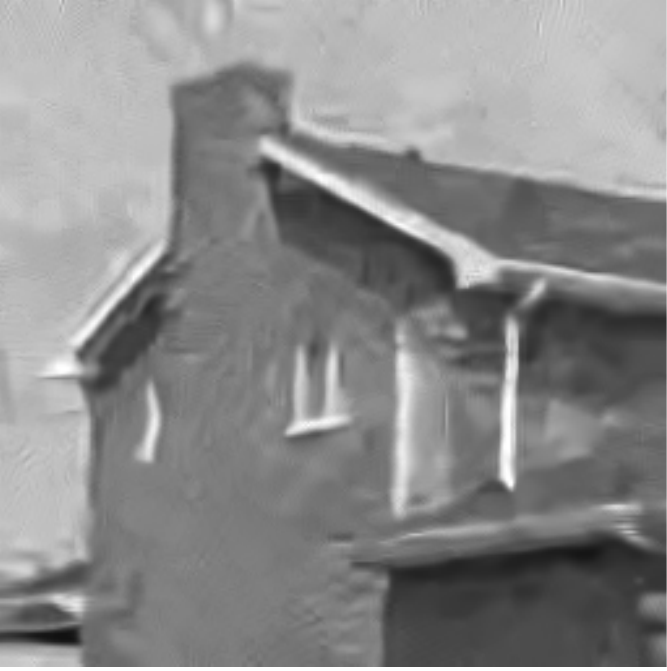}
}
\subfigure[{\scriptsize FoEPNR (\textbf{26.03}/\textbf{0.77})}]{
\centering
\includegraphics[width=0.22\textwidth]{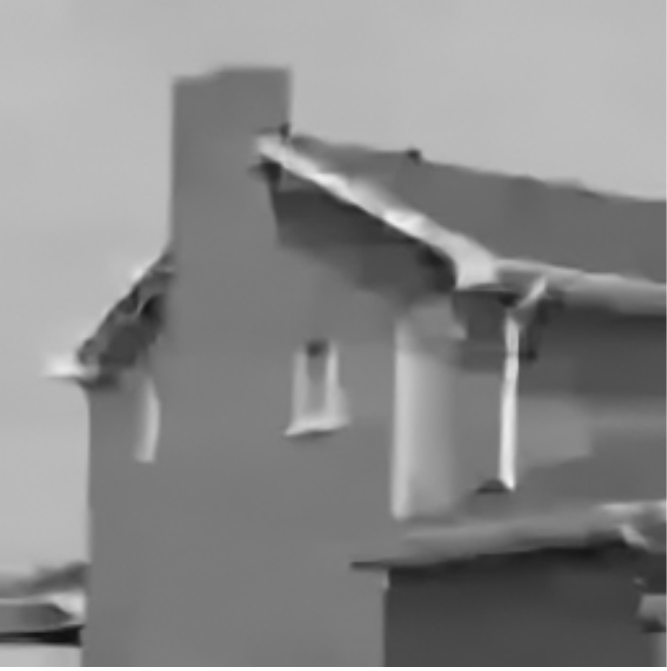}
}\\
\subfigure[{\sanhao Noisy image. Peak=4}]{
\centering
\includegraphics[width=0.22\textwidth]{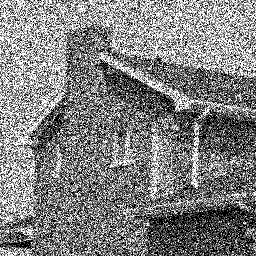}
}
\subfigure[{\sanhao NLSPCAbin (20.92/0.64)}]{
\centering
\includegraphics[width=0.22\textwidth]{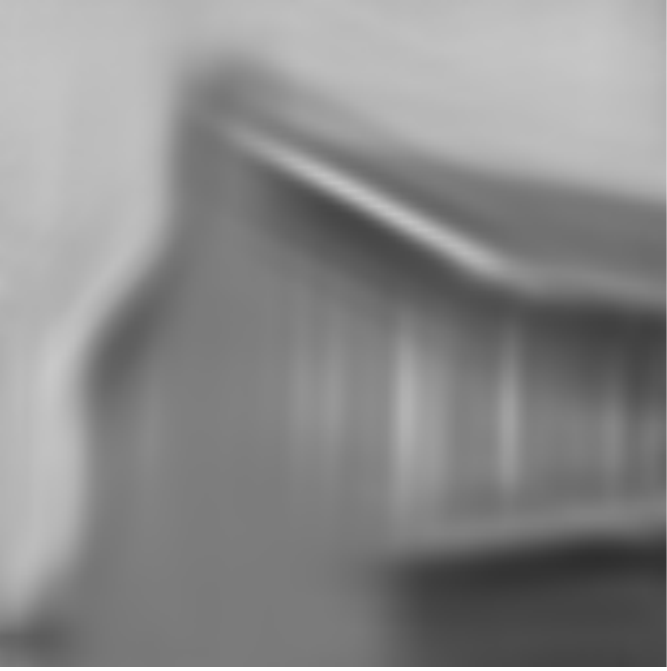}
}
\subfigure[{\sanhao BM3Dbin (25.08/0.74)}]{
\centering
\includegraphics[width=0.22\textwidth]{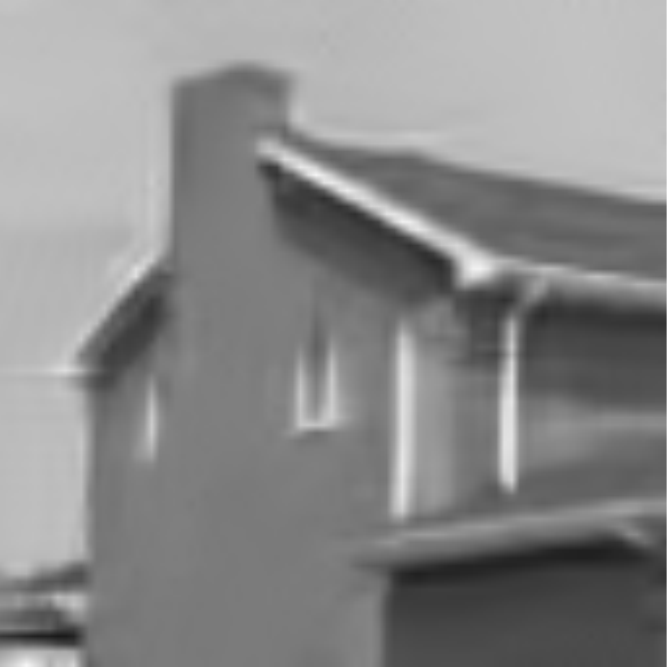}
}
\subfigure[{\sanhao FoEPNRbin (25.08/0.74)}]{
\centering
\includegraphics[width=0.22\textwidth]{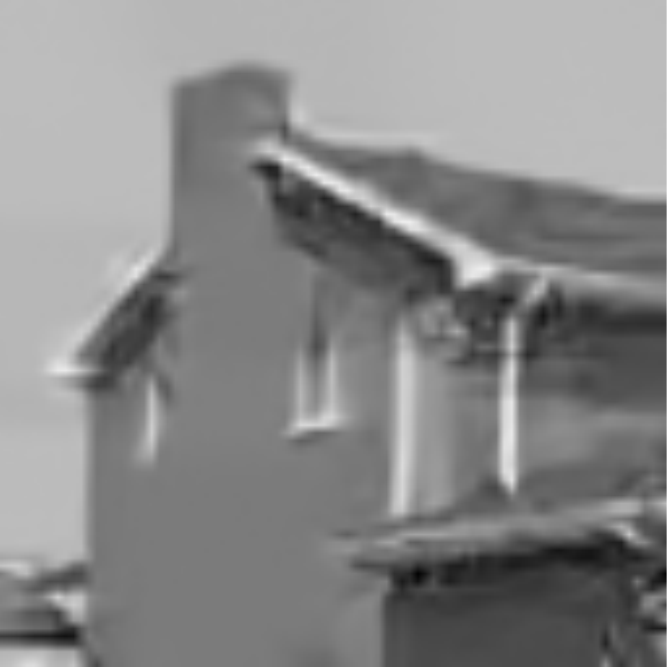}
}
\caption{Denoising of $house$ image with peak=4. The results are reported by PSNR/MSSIM index. Best results are marked.}
\label{peak4}
\end{figure*}

In Fig.~\ref{peak1}-Fig.~\ref{peak4}, the recovered results are reported for peak=1, 2 and 4 respectively. It can be observed that FoEPNR and BM3D perform best on detail preservation, and achieve evidently better results in term of PSNR/MSSIM index. Note that, the nonlocal technique BM3D is affected by the so-called ghost artifacts, structured signal-like patches that appear in homogeneous areas of the image, originated by random noise and reinforced through the patch selection process. The essential idea of BM3D is that only a few pixels with the most similar context are selected for the estimation. As a consequence, the selection process is inevitably influenced by the noise itself, especially in flat areas of the image, which can be dangerously self-referential.

The recovery error in terms of PSNR (in dB) and MSSIM are summarized in Table~\ref{resultshow}. Comparing the
indexes in Table~\ref{resultshow} and the denoising results in the present figures, we can
safely draw the following conclusions.
\begin{enumerate}
\item [$\mathrm{1)}$]For $\mathrm{peak} \leq 4$ where the Anscombe transform is less effective, the best indexes are guaranteed by our proposed FoE-based methods. The proposed methods employ a specialized FoE model trained for Poisson denoising problem and adaptive data terms to accommodate different Poisson noise levels, whereby they gain better results for the case of low peak value ($\leq 4$). As the peak value increases, the accuracy of the Anscombe transform increases, leading to the performance improvement of BM3D-based Poisson denoising algorithm. Note that by looking at the overall performance, our proposed variational model and the BM3D-based method behave similarly.
\item [$\mathrm{2)}$]In the rather low noise level ($\mathrm{peak}<1$), the introduction of the binning technique yields a significant performance enhancement. This is reasonable because when the noise level is extremely low, the operation of aggregating the noisy Poisson pixels into small bins results in a smaller Poisson image with lower resolution but higher counts per pixel. Therefore, one obvious advantage of the binning is that it can significantly reduces computation burden. However, as the peak raises, the binning becomes less effective. The reason is that if the peak value is not very low again, there is no need for the noisy image to suppress its noise level using binning at the loss of resolution reduction which will weaken or even eliminate many image details.
\item [$\mathrm{3)}$]In terms of MSSIM, the best overall performance is provided by the proposed methods (FoEPNR and FoEPNRbin) and BM3D-based method with the former slightly better, indicating that our method is more powerful in geometry-preserving, which can noticeably be visually perceived in Fig.~\ref{peak0.2}.
\item [$\mathrm{4)}$]In the visual quality, the typical structured artifacts encountered with the BM3D-based algorithm do not appear when the proposed methods are used. Meanwhile, we found that our method introduces block-type artifacts for the case of low peak values, e.g., $\mathrm{peak}=1$ in Fig.~\ref{peak1}(d). The main reason is that our method is a local model, which becomes less effective to infer the underlying structure solely from the local neighborhoods, if the input image is too noisy.
\end{enumerate}

For better describing the typical artifacts introduced by the proposed local-based methods, we briefly analyze the denoising results within the red rectangles of Fig.~\ref{peak0.2}(e) and Fig.~\ref{peak1}(e). By observation on the image structure within the red rectangle of Fig.~\ref{peak1}(e), it is understandable for FoEPNR to recover the result in Fig.~\ref{peak1}(d) which seems more similar to the underlying feature in the noisy image than Fig.~\ref{peak1}(c). However, the recovered result of BM3D-based method within the red rectangle is smoother and closer to the original clean image than that of FoEPNR. A similar phenomenon can also be observed in Fig.~\ref{peak0.2} where FoEPNR produces more serious artifact than BM3D-based method does. As mentioned above, for too noisy input images, the local method
becomes less effective. Therefore, some incorrect features caused by noise will disturb the denoised result of FoEPNR. Although this uncorrect feature is possibly not quite obvious, the trained FoE filters will still extract it and exhibit this feature apparently. In this case, the advantage of non-locality comes through by selecting several most similar context for estimating the underlying image intensity. That is to say, the denoised result at some location keep consistent with its similar contexts' information. However, if the uncorrect feature is apparent enough, both the nonlocal and local methods will extract it, as analyzed on Fig.~\ref{peak0.2}(c) and (d).

It is worthy noting that, in the aforementioned experiments we employ the case of one fixed noise realization because we want to provide determinant and easily reproducible results for a direct comparison between different algorithms. For a comprehensive evaluation, we also take 100 trials to compare the performances of the proposed method and BM3D-based method on image $Cameraman$ with many noise realizations. The peak value is set as 1. The average PSNR/MSSIM values are presented in Table~\ref{runtime100} from which we can see that the difference of the average PSNR/MSSIM using 100 different noise realizations is similar to that of PSNR/MSSIM obtained by the fixed pseudo-random noise realization adopted in our study. Moreover, for 100 repetitions of Table~\ref{resultshow}, the total time consumed by NLSPCA (referring to Table~\ref{runtime}) is about 1490 hours. Therefore, we have not presented the results based on many noise realization in consideration of time consumption. Nevertheless, the results of one fixed noise realization can already elucidate the difference of the test algorithms.

\begin{table}[t!]
\begin{center}
\begin{tabular}{c|c}
\cline{1-2}
BM3D & FoEPNR\\
\hline\hline
20.40/0.60 & 20.85/0.63\\
\cline{1-2}
20.36/0.61 & 20.80/0.64\\
\cline{1-2}
\end{tabular}
\end{center}
\caption{PSNR/MSSIM values. The first line of the table is obtained using one particular noise realization. The second line is obtained by averaging the PSNR/MSSIM values of 100 trials where each trial takes one different noise realization.}
\label{runtime100}
\end{table}

\subsection{Run time}
\begin{table}[t!]
\begin{center}
\begin{tabular}{r|c|c|c}
\cline{1-4}
& BM3D & NLSPCA & FoEPNR\\
\hline\hline
$256 \times 256$ & 1.38 & 367.9 &27.31 (\textbf{0.20})\\
$512 \times 512$ & 4.6 & 1122.1 &52.75 (\textbf{0.63})\\
\cline{1-4}
\end{tabular}
\end{center}
\caption{Typical run time (in second) of the Poisson denoising methods for images with two different dimensions.
The CPU computation time is evaluated on Intel CPU X5675, 3.07GHz.
The highlighted number is the run time of GPU implementation based on NVIDIA Geforce GTX 780Ti.}
\label{runtime}
\end{table}

It is worthwhile to note that our model is a local model with simple structure, as it merely contains
convolution of linear filters with an image, and therefore quite differs from the non-local models
(e.g., the BM3D-based Poisson denoising method). Our approach corresponds to a non-convex minimization problem, which is
efficiently solved by an iterative algorithm - iPiano. Each iteration consists of $48 \times 2 = 96$ convolution operations of filters
of size $7 \times 7$ and a few additional pixel-wise calculations. The overall required iterations vary for different noise levels.
For cases of relatively low peak, e.g.,
$\text{peak} = 4$, it generally consumes 150 iterations; while for cases of relatively high peak, e.g.,
$\text{peak} = 40$, usually 60 iterations is sufficient.
In Table \ref{runtime}, we report the typical run time of our model
for the images of two different dimensions exploited in this paper for the case of $\text{peak} = 4$.
We also present the run time of two competing algorithms for a comparison \footnote{
All the methods are run in Matlab with single-threaded computation for CPU implementation.
We only consider the version without binning technique.}.

Due to the structural simplicity of our model, it is well-suited to GPU parallel computation. We are able to implement our algorithm on
GPU with ease. It turns out that the GPU implementation
based on NVIDIA Geforce GTX 780Ti can accelerate the inference procedure significantly, as shown in Table \ref{runtime}.
\section{Conclusion}
We exploited variational models to incorporate the FoE prior, an widely used image prior/regularization model,
in the context of Poisson noise reduction. We went through a comprehensive study of various variants to evaluate
the corresponding performance. Finally, we arrived at the best variant based on (1) the Anscombe transformation, (2)
newly trained FoE image prior directly in the Anscombe transform domain and (3) an ad hoc setting of the date fidelity term.
This final version can lead to strongly competitive performance to state-of-the-art Poisson denoising methods, meanwhile, it
bears reasonable computation efficiency due to the usage of the iPiano algorithm. In summary,
we have demonstrated that a simple local model can also achieve state-of-the-art Poisson denoising performance, if
with appropriate modeling and training.

In our current work, we conducted our training experiments using the images from the BSDS300 image
segmentation database~\cite{amfm2011} which is targeted for natural images in a general sense.
However, the Poisson noise often arises in applications such as
biomedical imaging, fluorescence microscopy and astronomy imaging. In these typical applications, training specialized FoE prior model for a specific task bears the potential to
improve the current results. This is subject to our future research work.




{
\small
\bibliographystyle{plain}
\bibliography{references}
}

\end{document}